\documentclass[journal]{IEEEtran}
\usepackage[caption=false,font=normalsize,labelfont=sf,textfont=sf]{subfig}
\usepackage{stfloats}
\usepackage[pdftex]{graphicx}
\usepackage{amsmath}
\usepackage{amssymb}
\usepackage{textcomp,booktabs}
\usepackage[usenames]{color}
\usepackage{wrapfig}
\usepackage{colortbl,booktabs}
\usepackage{verbatim}

\usepackage{epstopdf}
\usepackage{multirow}
\usepackage{longtable}
\usepackage{rotating}

\begin{document}
%

\title{A Plug-and-play Scheme to Adapt Image Saliency Deep Model for Video Data}


\author{Yunxiao Li$^1$
~~~~~~Shuai Li$^1$ ~~~~~~Chenglizhao
Chen$^{1,2*}$\thanks{Corresponding author: Chenglizhao Chen,
cclz123@163.com.}  ~~~~~~Aimin Hao$^{1,3}$  ~~~~~~Hong Qin$^4$ \\ ~~~~~~~~~$^1$State
Key Laboratory of Virtual Reality Technology and Systems, Beihang University\\
$^2$Qingdao University~~~~~~~~~~~~~~$^3$Peng Cheng Laboratory~~~~~~~~~~~~~~$^4$Stony Brook University %
\\ Code\&Data: {https://github.com/YunX17/AdaptSaliency}


}

\markboth{IEEE Transactions on Circuits and Systems for Video Technology, VOL.XX, NO.XX, XXX.XXXX}%
{Shell \MakeLowercase{\textit{et al.}}: Bare Demo of IEEEtran.cls for Journals}

\maketitle

\IEEEtitleabstractindextext{
\begin{abstract}
With the rapid development of deep learning techniques, image saliency deep models trained solely by spatial information have occasionally achieved detection performance for video data comparable to that of the models trained by both spatial and temporal information.
However, due to the lesser consideration of temporal information, the image saliency deep models may become fragile in the  video sequences dominated by temporal information.
Thus, the most recent video saliency detection approaches have adopted the network architecture starting with a spatial deep model that is followed by an elaborately designed temporal deep model.
However, such methods easily encounter the performance bottleneck arising from the single stream learning methodology, so the overall detection performance is largely determined by the spatial deep model.
In sharp contrast to the current mainstream methods, this paper proposes a novel plug-and-play scheme to weakly retrain a pretrained image saliency deep model for video data by using the newly sensed and coded temporal information.
Thus, the retrained image saliency deep model will be able to maintain temporal saliency awareness, achieving much improved detection performance.
Moreover, our method is simple yet effective for adapting any off-the-shelf pre-trained image saliency deep model to obtain high-quality video saliency detection. Additionally, both the data and source code of our method are publicly available.
\end{abstract}


}
\maketitle
\IEEEdisplaynontitleabstractindextext
\IEEEpeerreviewmaketitle


\section{Introduction and Motivation}

\IEEEPARstart{I}MAGE saliency detection aims at locating the most eye-catching regions in a given scene~\cite{TIP15:2015,OurTIP15}.
As a pre-processing tool, its subsequent applications frequently include various computer vision applications, e.g., adaptive image retargeting~\cite{fang2012TIP},
image compression~\cite{TMM_li2017closed},
object tracking~\cite{PR_chen2015real}, and
video surveillance~\cite{CC2019CVPR,ChenPR16}.

Since the human visual system is extremely sensitive to the distinct movement patterns~\cite{TIP17chen2017video,OurTIP19,CC2020TIP}, the solely spatial-information-trained image saliency deep models may become fragile in video data whose saliency should be simultaneously determined by both spatial and temporal information~\cite{CC_TMM_2018}.
Thus, we may suppose that the current video saliency methods~\cite{li2018flow,songECCV2018pyramid,zhou2018video} that make full use of both the spatial and temporal information should significantly outperform the image saliency deep models~\cite{CMMTPAMI2019resDSS,chen2018eccvRAS,hu2018recurrent,TIP_li2016visual}.
However, more often than not, the opposite result is obtained due to the variable nature of video data.
For example, a salient object may occasionally be static for a long period showing no movement~\cite{SPL_chen2018novel} so that the spatial information may become the only saliency cue and, as a result, the spatiotemporal video saliency models cannot perform well in such case.
Thus, the overall performances of the up-to-date image saliency models~\cite{wangCVPR2019salient,fengCVPR2019attentive,wuCVPR2019cascaded,liuCVPR2019simple} are occasionally comparable to that of the state-of-the-art video saliency models.

\begin{figure}[!t]
	\begin{center}
		\includegraphics[width=1\linewidth]{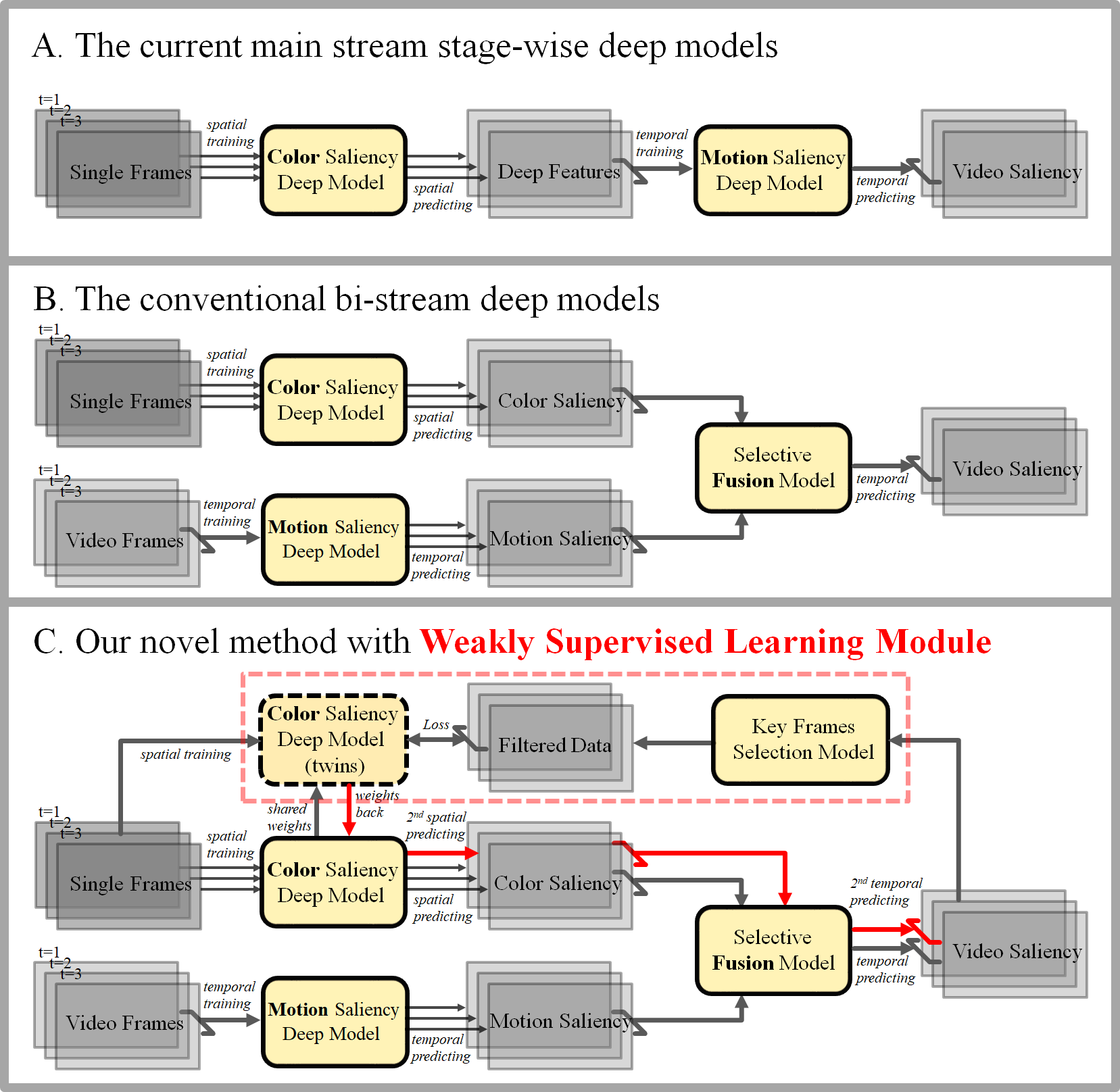}
	\end{center}
	\vspace{-0.4cm}
	\caption{Differences between our novel method and the conventional methods.}
	\label{fig:MotivationDemo}
	\vspace{-0.3cm}
\end{figure}

In fact, there are two types of mainstream network architectures that are prevalent in the video saliency detection field, i.e., the early bi-stream network architecture (Fig.~\ref{fig:MotivationDemo}-B) and the current single-stream network architecture (Fig.~\ref{fig:MotivationDemo}-A).
The conventional bi-stream network architecture consists of two independent sub-branches, of which one aims to conduct color saliency estimation from the spatial domain and the other  attempts to reveal motion saliency clues over the temporal scale.
Then, a fusion module is latterly applied to achieve a complementary status between the deep features of its precedent sub-branch, achieving spatial-temporal video saliency detection.
However, the overall performance of the methods based on the bi-stream architecture is strongly limited by the temporal sub-branch because it is much more difficult to conduct temporal saliency estimation directly from the video frames with an extremely large problem domain.

As a result, the most recent video saliency work~\cite{fanCVPR2019shifting} has adopted the single stream network architecture (Fig.~\ref{fig:MotivationDemo}-A) that performs the spatial-temporal saliency estimation within the coarse-to-fine manner; i.e., the preceding color saliency deep model aims to coarsely locate the salient regions in the spatial domain, while its subsequent motion saliency deep model attempts to finely filter the non-salient nearby surroundings over the temporal perspective.
Because the motion saliency deep model takes the output of its preceding color saliency deep model as the input (i.e., the spatial saliency deep features), its temporal saliency learning procedure can easily converge within a much simple problem domain.
Nevertheless, the single stream network architecture also encounters a chicken-and-egg problem resulting in a performance bottleneck; i.e., the performance of the motion saliency deep model is heavily dependent on its preceding color deep saliency model, however, the deep features provided by color deep model generally show limited performance due to its lesser consideration of the temporal information.

To address the above-mentioned problems, in this paper, we attempt to follow the conventional bi-stream network architecture and devise a novel weakly supervised learning scheme to break the performance bottleneck of the temporal sub-branch.
We have shown the overall methodology of our method in Fig.~\ref{fig:MotivationDemo}-C, and the key ideas of our method include the following 2 aspects:\\
1) We have devised a novel supervised fine-tune scheme that uses tiny amounts of newly sensed and coded temporal information to rapidly adapt any off-the-shelf color saliency deep model for a high-quality temporal saliency estimation;\\
2) Based on the saliency detection results of the bi-stream network, we attempt to rapidly identify video frames with high quality video saliency detections that will be subsequently used as the pseudo ground truth for the weakly supervised learning, enabling the color sub-branch to perform saliency estimation over both spatial and temporal domains.

In particular, it is important to mention that the performance of our method can be easily improved further when we adopt a much stronger pre-trained color saliency deep model as the color sub-branch.
Thus, with the rapid development of the color saliency deep learning techniques, our method will eventually be able to outperform the conventional video saliency detection methods.

%
%

\section{Background and Related Works}
\subsection{Image Saliency Detection Methods}
The conventional image saliency detection methods frequently adopt multiple discriminative hand-crafted saliency cues to obtain a robust detection result, e.g., the most representative regional contrast saliency cue~\cite{Global_RC:2011} and the multi-level hierarchical saliency cue~\cite{yan2013hierarchical}.

After entering the deep learning era, the classic deep learning techniques, e.g., convolutional neural networks (CNNs) and full convolutional networks (FCNs), have been widely applied for image saliency detection.
Since the automatically computed deep features have much more discriminative semantic information, the deep learning based image saliency detection methods can significantly outperform the conventional hand-crafted methods.
Li et al.~\cite{TIP_li2016visual} propose to simultaneously make full use of both the CNN-based deep features and the hand-crafted low-level features, achieving much improved detection performance.
Furthermore, Li et al.~\cite{liTNNLS2018contrast} also propose a hybrid contrast-oriented deep neural networks method that takes full advantage of an attentional module to alleviate its computational burden.

The multi-level feature aggregation scheme has also been widely adopted in the image saliency detection field.
Hou et al.~\cite{CMMTPAMI2019resDSS} resort to the short connections to combine both deeper layers and shallower deep features in FCNs, which promotes the saliency detection accuracy effectively.
Although the method~\cite{CMMTPAMI2019resDSS} has achieved significant performance, it suffers from heavy computational burden.
Hence, Chen et al.~\cite{chen2018eccvRAS} propose the residual learning scheme to further refine the side layers' deep features, shrinking the parameters of its convolutional layers effectively.
Hu et al.~\cite{hu2018recurrent} equip their network with recurrently aggregated deep features captured in the different layers of Fully Convolutional Networks (FCNs) to more accurately detect salient objects.

In~\cite{wuCVPR2019cascaded}, Wu et al. point out the wrong recognition toward the multi-level learning scheme; i.e., shallow deep features contribute less to the overall detection performance but have a higher computational cost.
Therefore, \cite{wuCVPR2019cascaded} propose to directly ignore these deep layers from the shallow layers to further increase the computational speed.

Most recently, Liu et al.~\cite{liuCVPR2019simple} propose a global guidance module from the bottom-up pathway and a feature aggregation module from the top-down pathway, fusing them in a multi-scale manner to enrich the details of the saliency detection.
Wang et al.~\cite{wangCVPR2019salient} propose a pyramid attention structure for salient object detection, enhancing the representation ability of its deep features effectively.
Moreover, Feng et al.~\cite{fengCVPR2019attentive} further utilize an attentive feedback network with boundary-enhanced loss to further sharpen the detected salient object boundaries.

\subsection{Hand-crafted Video Saliency Detection Methods}
The conventional hand-crafted video saliency detection methods~\cite{Zhou2018vcir,fang2014tip,TIP15:2015,xiTIP2016salient,cheny2018tip} usually utilize low-level cues to construct an elaborately designed optimization graph for maintaining saliency spatiotemporal coherency.
Wang et al.~\cite{TIP15:2015} propose a spatial-temporal energy function with a gradient flow field to obtain spatiotemporally consistent saliency maps that are then further improved by using the newly designed appearance model and location model~\cite{wang2015saliency}.
Similarly, by using multiple low-level features, Kim et al.~\cite{Kim:2015} construct a probability framework consisting of spatial transition matrices with temporal restarting distribution in order to evaluate the video saliency via random walk with restart.
Xi et al.~\cite{xiTIP2016salient} propose to feed multiple newly revealed spatiotemporal background priors into a dual-graph network, achieving much improved detection performance.

Unlike for the above-mentioned graph optimization based methods, the spatiotemporal coherency can also be sustained within a batch-wise manner.
Li et al.~\cite{CVIU_li2015kernel} propose a kernel regression that includes three entity-models to exploit the spatiotemporal coherency of attentional regions.
Chen et al.~\cite{TIP17chen2017video} propose to conduct low-rank guided batch-wise regional alignments, fulfilling the spatiotemporal coherency by constraining the saliency similarly between any aligned regional pairs.
Further, Chen et al.~\cite{CC_TMM_2018} introduce  a bi-level feature learning scheme to further expand the spatiotemporal coherency sensing scope from the long-term perspective.
Liu et al.~\cite{liu2017saliencySGSP} perform temporal and spatial propagation via similarity matrices, obtaining saliency maps with strong spatiotemporal coherent.
Zhou et al.~\cite{zhouTMM2018improving} employ localized estimation models with spatiotemporal refinement mechanism to further improve the detection performance.
Guo et al.~\cite{guoTCB2017video} propose to integrate the conventional motion saliency cues into the object proposals.
The temporal saliency cues are much more stable than the spatial saliency cues.
Therefore, Guo et al.~\cite{guoTCSVT2019motion} develop a rapid video saliency detection method using the
principal motion vector and an appearance cue to achieve
temporal consistency.

\subsection{Video Saliency Deep Models}
As previously mentioned, almost all of the deep learning based video saliency object detection methods can be divided into two types, namely, the bi-stream and the single stream methods.
The most representative bi-stream method~\cite{wang2018video} proposes a two-stream network, feeding static saliency into the module for dynamic saliency, obtaining saliency with a lower computational load.
Addressing the same issue, Li et al.~\cite{li2018flow} design a universal framework to increase the temporal coherence of the deep feature representation with ConvLSTM, achieving high speed.
After attaching a fast-moving object edge map to the bi-stream based network, Sun et al.~\cite{sun2018sg} combine memory information to achieve robust detection.
To deal with the lack of manually labeled data, Tang et al.~\cite{tang2018weakly} train two cascade fully convolutional networks in a weakly supervised manner to predict saliency via both spatial and temporal cues.
Le et al.~\cite{le2018video} propose to detect salient foregrounds by using conventional convolution in spatial branch and 3D convolution in the temporal branch over regions and consecutive frames, respectively.
Similar to the method in \cite{le2018video}, Fang et al.~\cite{fang2018deep3dsaliency} apply STSM and SSAM to estimate saliency over the time axis and spatial coordinate, respectively.
And Wen et al.~\cite{wen2019deep} generate saliency via fusing multi-level deep features extracted by a symmetrical CNN composed of spatial and temporal branches.

The methods recently described in ~\cite{li2019motion} and ~\cite{xu2019spatiotemporal} follow the same bi-stream approach, introducing two similar yet effective architectures that are based on conventional bi-stream, employing two sub-networks for detecting saliency in still images and temporal data while using motion sub-network to enhance the sub-network for still images.
In particular, Yan et al.~\cite{yan2019semi} present an effective spatial refinement network and then used a recurrent module to obtain both accurate contrast inference and coherence enhancement.

The bi-stream methods frequently lack relatively adequate yet gratifying ability to accomplish temporal saliency estimation in the video frames.
Single stream methods shrink the problem domain via considering the output of the color saliency deep model as the input for the latter branch to estimate temporal saliency, achieving the best performance among the currently available methods.
To the best of our knowledge, there are only several methods following the single stream approach.
Specifically, observing the burden of considering multi-scale features in Region-based CNN, Song et al.~\cite{songECCV2018pyramid} use ConvLSTM with pyramid dilated convolution architecture to obtain consistency in the large space-time margin.
Based on the method  of~\cite{songECCV2018pyramid}, Fan et al.~\cite{fanCVPR2019shifting} integrate a saliency-shift loss guided attention mechanism to strengthen the discrimination of the ConvLSTM network.

From the perspective of conventional LSTMs, the internal state of each memory cell contains the accumulated information about the spatiotemporal structure, while it may fail to capture these salient motions, because the gate units do not explicitly consider the impact of dynamic structures present in input sequences.
To solve this limitation, Veeriah et al.~\cite{Veeriah2015Differential} introduce a differential RNN (dRNN) model that integrates the Derivative of States (DoS) into the conventional RNN, aiming to quantifies the change of information at each time thereby learning the evolution of action states.
It is worthy mentioning that the concept of ``salient motion'' used in the dRNN usually relates to those motions that make the current action more discriminative than others towards the action recognition task.
In fact, the key idea of our work is partially similar to the dRNN, which attempts to resort those frames containing salient motions to impact the upcoming new round of network training.
However, there exists one critical aspect making the meaning of ``salient motion'' in our method different to that in the dRNN, i.e., the DoS adopted in the dRNN can only indicate whether the current frame contains `salient motion'---it may be the partial movements, belonging to different objects or even dynamic backgrounds.
Though such kind of salient motions can benefit the action recognition task, it may be not suitable for the video salient object detection because of the non-quality-aware nature of the DoS.
In sharp contrast, our method resorts the consistency between color saliency and motion saliency to measure the quality of frames, and only those frames containing high-quality motions will be used to impact the upcoming re-learning process, which is more suitable for the video salient object task.

Specifically, our idea is partially similar to~\cite{griffin2019bubblenets,li2019effective} in the video object segmentation task, of which the key idea of paper~\cite{griffin2019bubblenets,li2019effective} is to utilize keyframe strategy for fine-tuning.
However, our method is different from them. BubbleNets~\cite{griffin2019bubblenets} learns to sort frames via a performance-based loss function and all the data for training the network derives from annotated dataset. After that, the method employs one selected keyframe with its accurate groundtruth for fine-tuning. Therefore, this method is a supervised method.
On the contrary, our method is a semi-supervised one since we only need pseudo groundtruth where we directly employ those low-level saliency maps instead of accurate annotated data in our online training.
As for the other paper, Li et al.~\cite{li2019effective} attempt to re-train the model with the first frame and the selected keyframes which are obtained via several metrics of segmentation quality, e.g., average region number, temporal pixel change rate and segmentation compactness. On the one hand, the main task is different from ours because the paper~\cite{li2019effective} needs to know an accurate groundtruth of the first frame in advance.
On the other hand, all the metrics of segmentation quality adopted in the paper~\cite{li2019effective} are based on the conventional hand-crafted rationale. In sharp contrast, our novel training scheme takes advantage of both deep model based motion saliency extracted from color-coded optical flow and color saliency extracted from original frame. In many cases, our method is more robust than simply applying hand-crafted features. In addition, our method has a relatively higher computing efficiency.

\begin{figure}[!t]
	\begin{center}
		\includegraphics[width=1\linewidth]{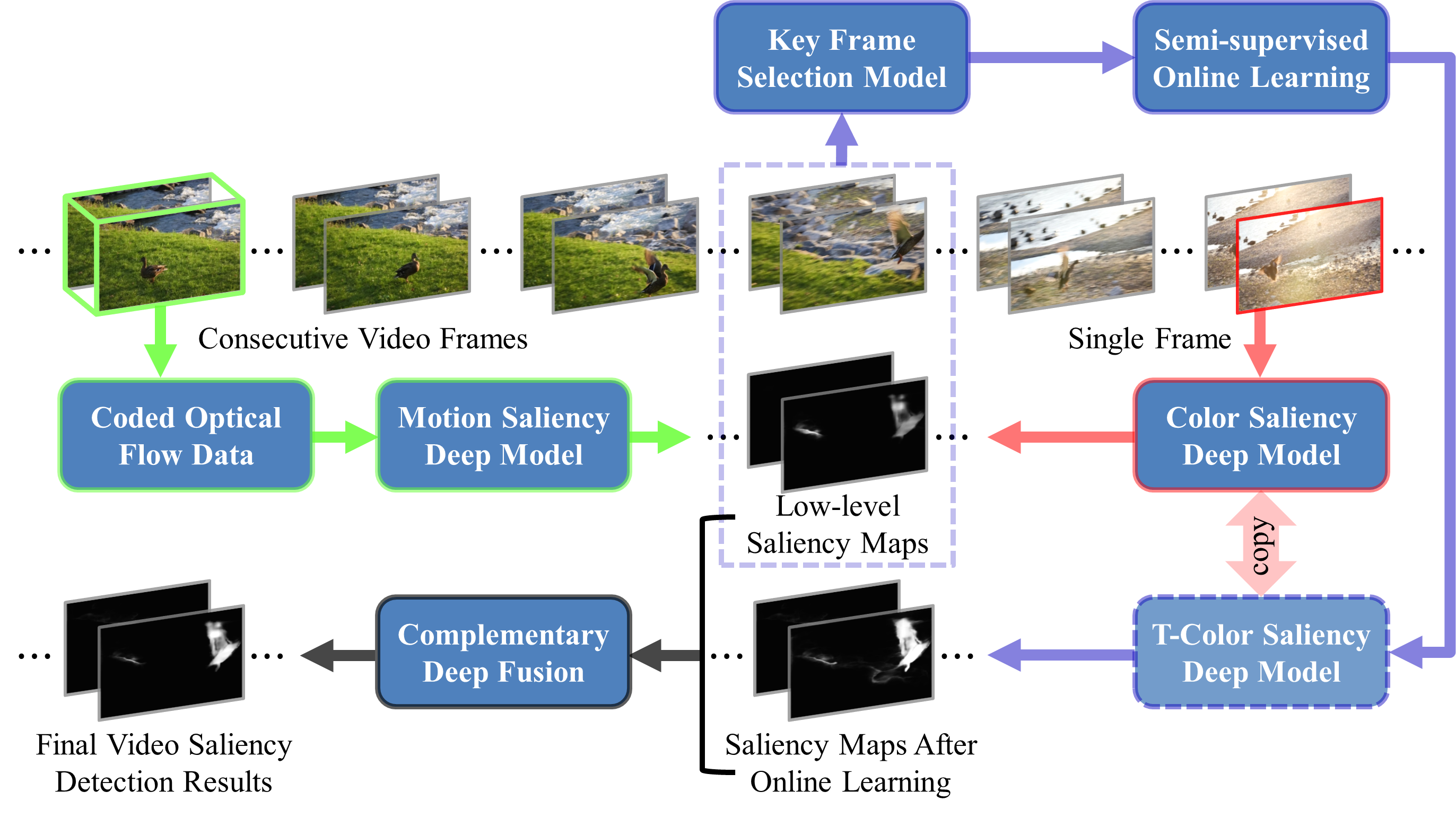}
	\end{center}
	\vspace{-0.4cm}
	\caption{Architectural overview of our video saliency detection method in which the red arrows denote the color-related data flows, the green arrows denote the motion-related data flows, the blue arrows denote the data flows of our weakly supervised online learning scheme, and the black arrows denote the final bi-stream spatiotemporal saliency fusion.}
	\label{fig:Pipeline}
	\vspace{-0.4cm}
\end{figure}

\section{Method Overview}
\label{sec:Method Overview}
As shown in Fig.~\ref{fig:Pipeline}, our method consists of
6 steps:
STEP 1. Color-coded optical flow computation;
STEP 2. Motion saliency computation;
STEP 3. Color saliency computation;
STEP 4. Low-level saliency computation;
STEP 5. Perform our keyframe strategy to locate the informative video frames and an online learning is proposed in the weakly supervised manner;
STEP 6. Complementary fusion between the pre-computed low-level saliency and the newly computed saliency maps after using the weakly supervised online learning, ensuring the temporal consistency in the long-term manner to improve the overall detection performance.

Given an input video, we first compute its color-coded optical flow. According to the optical flow data, motion saliency maps are generated via our newly trained high performance motion saliency deep model, as will be detailed in Sec.~\ref{sec:MotionSaliencyLearning}.
Furthermore, we use the pre-trained color deep saliency model to initially conduct color saliency estimation and then fuse it with the motion saliency map as the low-level saliency map, see details in Sec.~\ref{sec:Low-levelSaliencyComputation}.

Based on the computed low-level saliency maps, we use the newly designed keyframe strategy (Sec.~\ref{sec:Low-level Saliency Guided Key Frame Selection}) to locate the frames with relatively high-quality low-level saliency predictions that will subsequently be applied as the pseudo training ground-truth. Then, we adopt the newly self-paced online learning scheme (Sec.~\ref{sec:Self-paced Online Learning}) to re-train the color deep model, in which the re-trained color deep model can simultaneously conduct saliency estimation from both the spatial and temporal perspectives.
Finally, we fuse the original low-level saliency maps with the saliency predictions computed by the newly re-trained color deep model, achieving an optimal fusion status with much improved detection performance.

\section{Model Adaption}
\label{sec:IntraFrameSaliency}

\subsection{Color Saliency Deep Model Preliminaries}
\label{sec:ColorModelPreliminaries}
Almost all of the current mainstream FCN-based image saliency deep models~\cite{liTNNLS2018contrast,wang2016saliency} have enabled the end-to-end saliency detection, requiring much less computational cost than the conventional CNN-based methods.
Considering feature map $\textbf{X}$ and feature map $\textbf{X}'$ after convolution operation, the convolutional operator can be formulated as given by Eq.~\ref{eq:convolution}.
\begin{equation}
\label{eq:convolution}
\textbf{X}' = \textbf{W}*\textbf{X} + \textbf{b},
\end{equation}
where $\textbf{W}$ and $\textbf{b}$ denote kernel and bias, respectively, in which the convolutional operator is frequently applied to down-sample its input.

Although the down-sampled feature maps are valuable for coarsely locating the salient objects, they tends to lose the tiny details, generating the detected saliency map with obscured object boundary.
To alleviate this problem, the FCN-based state-of-the-art methods frequently adopt multiple de-convolution layers, and the overall forward propagation of the standard FCNs can be formulated as Eq.~\ref{eq:forwardpropagation}.
\begin{equation}
\label{eq:forwardpropagation}
\hat{\textbf{S}} = DeConv\Big (Conv(\textbf{I};\boldsymbol{\alpha});\boldsymbol{\beta}\Big ),
\end{equation}
where the function $DeConv(\cdot)$ denotes a series of de-convolutional operators to ensure the feature map of last layer with resolution identical to the input image $\textbf{I}$, $Conv(\cdot)$ denotes the convolutional operations in the adopted backbone network (e.g., VGG),
the symbol $\hat{\textbf{S}}$ denotes the final saliency prediction, and symbols $\boldsymbol{\alpha}$ and $\boldsymbol{\beta}$ represent all of the learned parameters of the convolution and de-convolution layers, respectively.

To measure and minimize the error in the training stage, the FCN-based methods commonly use the cross-entropy loss given by Eq.~\ref{eq:crossentropyloss}.
\begin{equation}
\label{eq:crossentropyloss}
\begin{split}
L(\boldsymbol{\alpha};\boldsymbol{\beta}) =& -\sum_{i,j} \textbf{G}_{i,j}\cdot log\Big(P(\textbf{G}_{i,j}=1|\boldsymbol{\alpha};\boldsymbol{\beta};\textbf{I})\Big)\\
&-\sum_{i,j}(1-\textbf{G}_{i,j})\cdot log\Big(P(\textbf{G}_{i,j}=0|\boldsymbol{\alpha};\boldsymbol{\beta};\textbf{I})\Big),
\end{split}
\end{equation}
where $\textbf{G}_{i,j}=1$ denotes foreground, and $\textbf{G}_{i,j}$ denotes the value of the ground-truth map at location $(i,j)$, e.g., $\textbf{G}_{i,j}=0$ denotes background;
$P$ denotes the probability of the final activation value at location $(i,j)$.

%

\begin{figure}[!t]
	\begin{center}
		\includegraphics[width=0.95\linewidth]{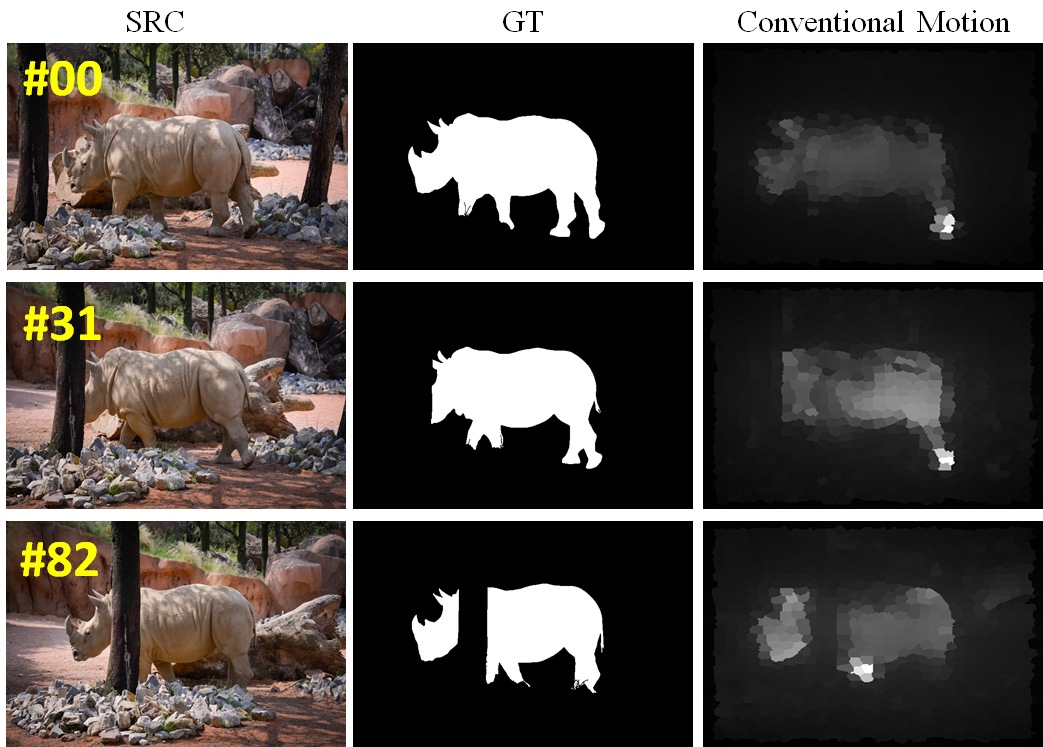}
	\end{center}
	\vspace{-0.4cm}
	\caption{An failure case demonstration that the conventional motion saliency (FL18~\cite{CC_TMM_2018}) toward the partial movements.}
	\label{fig:MotionDemo}
	\vspace{-0.3cm}
\end{figure}

\begin{figure*}[!b]
	\begin{center}
		\includegraphics[width=1.0\linewidth]{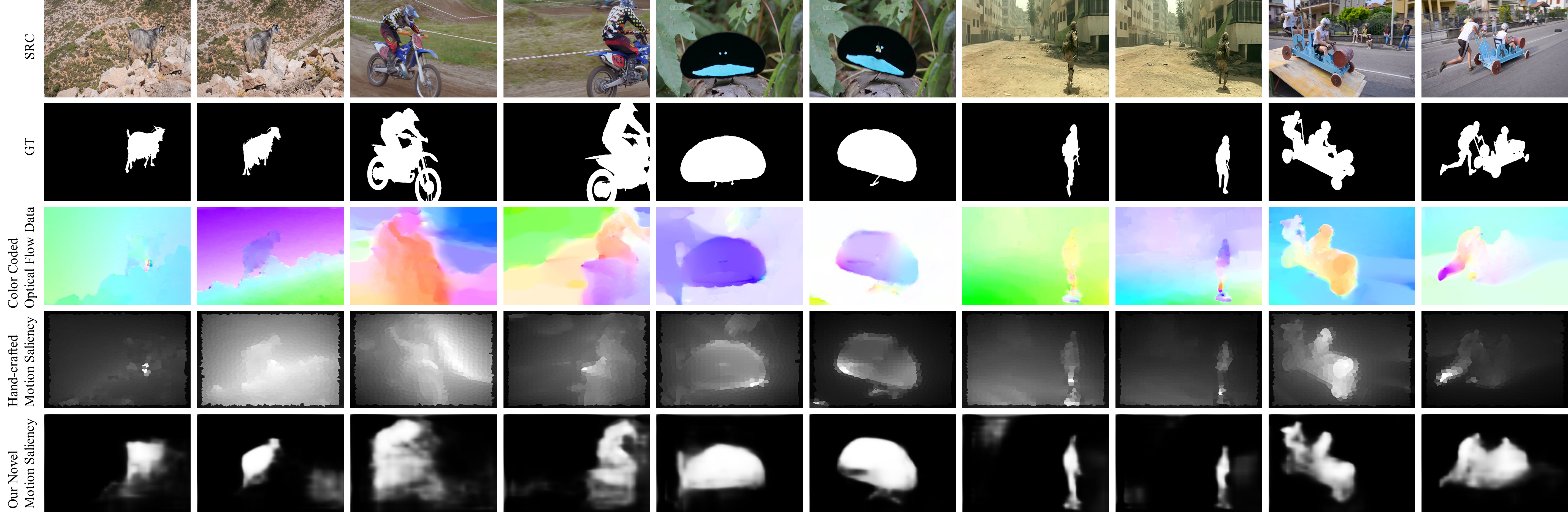}
	\end{center}
	\vspace{-0.4cm}
	\caption{Qualitative demonstrations of the performance improvement of our method, where GT denotes the saliency ground truth; we show the color-coded optical flow data in the 3-rd row; the motion saliency maps obtained by the hand-crafted method~\cite{CC_TMM_2018} and our method are demonstrated in 4-th and 5-th rows respectively.}
	\vspace{-0.4cm}
	\label{fig:motiondemo}
\end{figure*}

\subsection{Conventional Motion Saliency}
\label{sec:The Conventional Motion Saliency}
As one of the most important saliency cues in video data, the motions between two consecutive video frames can be easily sensed by the optical flow algorithm within a pixel-wise manner.
Given two consecutive video frames, the optical flow algorithm will output two directional (horizontal and vertical) gradient maps, i.e., ${VX}\in\mathbb{R}^{w\times h}$ and ${VY}\in\mathbb{R}^{w\times h}$, in which $w$ and $h$ denote the width and height, respectively, of the given image \textbf{I}.
In fact, while the heavy computational cost is the major performance bottleneck of the conventional optical flow algorithm~\cite{OpticalFlow:2009}, we may choose to use the deep-learning-based optical flow method (e.g., LiteFlowNet~\cite{Hui_CVPR_2018}) to achieve an extremely low computational cost at the expense of a slight performance degeneration.

Since the human visual system is extremely sensitive to the changes over the temporal direction, motion saliency (\textbf{MS}) can be easily inferred by conducting the pixel-wise or regional-wise contrast computation over the optical flow provided $VX$ and $VY$ as Eq.~\ref{eq:MotionSaliency}.
\begin{equation}
\label{eq:MotionSaliency}
\textbf{MS}_{i,j} = \sum_{u,v\in \Phi_{i,j}} \frac{||(VX_{i,j},VY_{i,j}),(VX_{u,v},VY_{u,v})||_2}{\omega\cdot||(i,j),(u,v)||_2},
\end{equation}
where $\omega$ is an empirically predefined weighting parameter, $||\cdot||_2$ denotes the L2 Euclidean distance, and
$\Phi_{i,j}$ denotes the region in the vicinity of pixel $(i,j)$.

In fact, while Eq.~\ref{eq:MotionSaliency} ensures that motion saliency computation is effective in most cases, it may still encounter cases with low-quality motion saliency that is mainly induced by either the anomaly object movements (e.g., the partial/intermittent movement in Fig.~\ref{fig:MotionDemo}) or the fast view scale/angle change (e.g., camera jitter); see the qualitative demonstrations in Fig.~\ref{fig:motiondemo}).

\subsection{A Novel Scheme to Adapt Color Saliency Deep Model for Motion Saliency Estimation}
\label{sec:MotionSaliencyLearning}
The deep-learning-based color saliency methods have received extensive research attentions with much significant performance improvements in the past five years.
However, the up-to-date motion saliency computations still follow either the hand-crafted contrast computation mentioned in Sec.~\ref{sec:The Conventional Motion Saliency} or the solely FCN-based simple networks~\cite{wang2018video} for which the core methodology has rarely benefited from the up-to-date color saliency related deep learning techniques.
Moreover, the current mainstream deep-learning-based color saliency methods can easily assign large saliency values to the regions that are perceptually distinct from their surroundings.

Thus, all of the above issues motivate us to explore a feasible approach for converting the off-the-shelf color saliency deep models for the high-performance motion saliency estimation.
In fact, the underlying rationale of saliency computation over temporal domain is in many ways identical to the classic color saliency computation, yet the difference between them is that the motion saliency is developed from the contrast computation over the optical flow spanned feature space rather than from the color-saliency-based spatial contrast.
Therefore, we propose to use the optical flow data to fine-tune a pre-trained color saliency deep model (we choose ResDSS~\cite{CMMTPAMI2019resDSS} in this paper) for motion saliency detection.

\begin{table}[!t]
	\caption{Quantitative comparison between different motion deep model training strategies.}
\resizebox{1\columnwidth}{!}{
	\begin{tabular}{c|c|c|c|c|c|c|c}
    \toprule[1pt]
		\multirow{2}[1]{*}{DataSet} & \multirow{2}[1]{*}{Metric}
		
		&
		\multicolumn{1}{c}{Ours} &
		\multicolumn{1}{c}{Ours} & \multicolumn{1}{c}{Ours} &
		\multicolumn{1}{c}{Ours} &
		\multicolumn{1}{c}{Ours} & Conventional
		\\

		&    & \multicolumn{1}{c}{$\rm ResDSS^+$}  &  \multicolumn{1}{c}{$\rm ResDSS^*$}
		& \multicolumn{1}{c}{$\rm ResDSS^{S}$}
		& \multicolumn{1}{c}{$\rm ResDSS^{P}$}
		&  \multicolumn{1}{c}{RADF} & \multicolumn{1}{c}{Motion (Eq.~\ref{eq:MotionSaliency})}
		\\
		
		\hline
		\multirow{3}[1]{*}{DAVIS-T~\cite{Perazzi2016benchmark}} & \multicolumn{1}{c|}{maxF} &  \multicolumn{1}{c}{{0.755}}  & \multicolumn{1}{c}{0.764} &
		\multicolumn{1}{c}{0.763} &
		\multicolumn{1}{c}{0.804} &
		 \multicolumn{1}{c}{{0.743}} & {0.645}
		\\
		
		& \multicolumn{1}{c|}{avgF} & \multicolumn{1}{c}{{0.657}} & \multicolumn{1}{c}{0.666} &
		\multicolumn{1}{c}{0.664} &
		\multicolumn{1}{c}{0.689} & \multicolumn{1}{c}{0.701} & {0.474}
		\\
		
		& \multicolumn{1}{c|}{MAE} & \multicolumn{1}{c}{{0.063}} & \multicolumn{1}{c}{0.079} &
		\multicolumn{1}{c}{0.080} &
		\multicolumn{1}{c}{0.065} & \multicolumn{1}{c}{0.056} & {0.176}		
		\\
	
    \toprule[1pt]
	\end{tabular}%
	}
    \vspace{-0.4cm}
	\label{tab:MotionComponent}%
\end{table}%

To handle the inconsistent data channel between optical flow data (i.e., 2-channel $VX,VY$) and original color image (i.e., 3-channel RGB), we follow the coding scheme mentioned in~\cite{baker2011database} to convert the 2-channel optical flow data into the 3-channel color-coded version, in which it uses the 55 pre-defined colors with different hue and saturation to represent the flow orientation and magnitude, respectively; see the color coded optical flow data in the 3rd row of Fig.~\ref{fig:motiondemo}.

Next, we will use the newly coded optical flow data to fine-tune the color saliency deep model, where the key steps can be summarized as follows:\\
1) we use the widely adopted training set (30 sequences) of the Davis dataset to train our motion model;\\
2) for each video frame in the adopted training set, we use the optical flow method~\cite{OpticalFlow:2009} to compute its optical flow data and then convert these data into 3-channel data using the aforementioned coding scheme;\\
3) we use the coded optical flow data to fine-tune the pre-trained ResDSS model with the learning rate of 1e-9 using the total loss function ($L_{total}$) given by Eq.~\ref{eq:totalcrossentropyloss}.
\vspace{-0.1cm}
\begin{equation}
\label{eq:totalcrossentropyloss}
\begin{split}
L_{total}&(\boldsymbol{\alpha};\boldsymbol{\beta}) = \\
&-\sum_{k}\sum_{i,j} \textbf{G}_{i,j} log\Big(P_k(\textbf{G}_{i,j}=1|\boldsymbol{\alpha}_k;\boldsymbol{\beta}_k;\textbf{F})\Big)\\
&-\sum_{k}\sum_{i,j}(1-\textbf{G}_{i,j}) log\Big(P_k(\textbf{G}_{i,j}=0|\boldsymbol{\alpha}_k;\boldsymbol{\beta}_k;\textbf{F})\Big),
\end{split}
\vspace{-0.1cm}
\end{equation}
where $\textbf{F}$ represents the input color-coded optical flow image, and $k$ denotes the index of the side-output layer or the fusion layer in ResDSS.
Thus, the final motion saliency $\hat{\textbf{MS}}$ can be formulated as $\sum_{k}\hat{\textbf{S}}_k/|k|$, where  the $\hat{\textbf{S}}_k$ denotes the $k$-th motion saliency prediction, and $|k|$ denotes the total number of the prediction maps.
The overall qualitative demonstration of the computed motion saliency maps is shown in Fig.~\ref{fig:motiondemo}.

Our motion deep model (i.e., the re-trained ResDSS) can be trained fast (within 2 hours) by using relatively little training data (only 2K).
Moreover, our method is flexible and can adapt any color saliency deep model for motion saliency detection.
In particular, it is important to mention that the motion saliency detection performance can easily be further improved if we adopt a much stronger pre-trained color saliency deep model; for example, we have re-trained the RADF model~\cite{hu2018recurrent} to achieve better motion saliency performance (see Ours RADF in Table~\ref{tab:MotionComponent}).

It is worthy mentioning that the overall performance of our method may vary with different optical flow sources, and we may well achieve better overall performance if we adopt more accurate optical flow method, such as the PWCNet~\cite{sun2018pwc}.
With regarding to this issue, please refer to the quantitative proofs in Table~\ref{tab:MotionComponent}, where the $\rm ResDSS^+$
represents the results using the conventional optical flow data, the $\rm ResDSS^*$ represents the re-trained motion model
using the optical flow data of the LiteFlowNet~\cite{Hui_CVPR_2018}, the $\rm ResDSS^{S}$ represents the re-trained motion model
using the optical flow data of the SPyNet~\cite{ranjan2017optical}, the $\rm ResDSS^{P}$ represents the re-trained motion model
using the optical flow data of the PWCNet~\cite{sun2018pwc}.

\subsection{Low-level Saliency Computation}
\label{sec:Low-levelSaliencyComputation}
We have obtained the newly learned motion saliency $\hat{\textbf{MS}}$ that can be used as the motion sub-branch in our bi-stream network mentioned in Sec.~\ref{sec:Method Overview}.
Moreover, in our bi-stream network, any off-the-shelf pre-trained color saliency deep model can be used as the color sub-branch and we represent its saliency prediction, namely color saliency map as $\hat{\textbf{CS}}$.
Thus far, we formulate the low-level saliency via the widely adopted multiplicative-based fusion as given by Eq.~\ref{eq:LowlevelSaliency}.

\begin{equation}
\label{eq:LowlevelSaliency}
\textbf{LS} = \hat{\textbf{MS}} \odot \hat{\textbf{CS}},
\end{equation}
where $\odot$ denotes the element-wise Hadamard product.
Since the fusion procedure has simultaneously considered both the spatial and temporal saliency cues, its overall performance can be superior to either the motion saliency or the color saliency, as quantitatively demonstrated in Sec.~\ref{sec:AdaptivenessAnalysis}.

\section{Weakly Supervised Online Learning}
\label{sec:weaklysupervisedtemporalconsistencylearning}

\subsection{Keyframe Strategy}
\label{sec:Low-level Saliency Guided Key Frame Selection}
Compared with the conventional contrast computation based motion saliency, our novel motion saliency deep model can output the motion saliency between the consecutive video frames correctly.
However, due to its limited sensing scope (only 2 frames) over the temporal scale, the motion saliency produced by our motion saliency deep model may be perceptually different from the real motion saliency.
Moreover, the sensing scope of the color branch in our bi-stream network is also limited within a single video frame, making the fused low-level saliency (\textbf{LS}) temporally inconsistent.
To solve this issues, here, we propose a novel weakly supervised online learning to adapt the color branch for a long-term spatiotemporal saliency detection.
It is also important to mention that we choose to leave the motion branch unchanged to avoid over-fitting.

\vspace{-0.4cm}
\begin{figure}[!h]
	\begin{center}
		\includegraphics[width=0.95\linewidth]{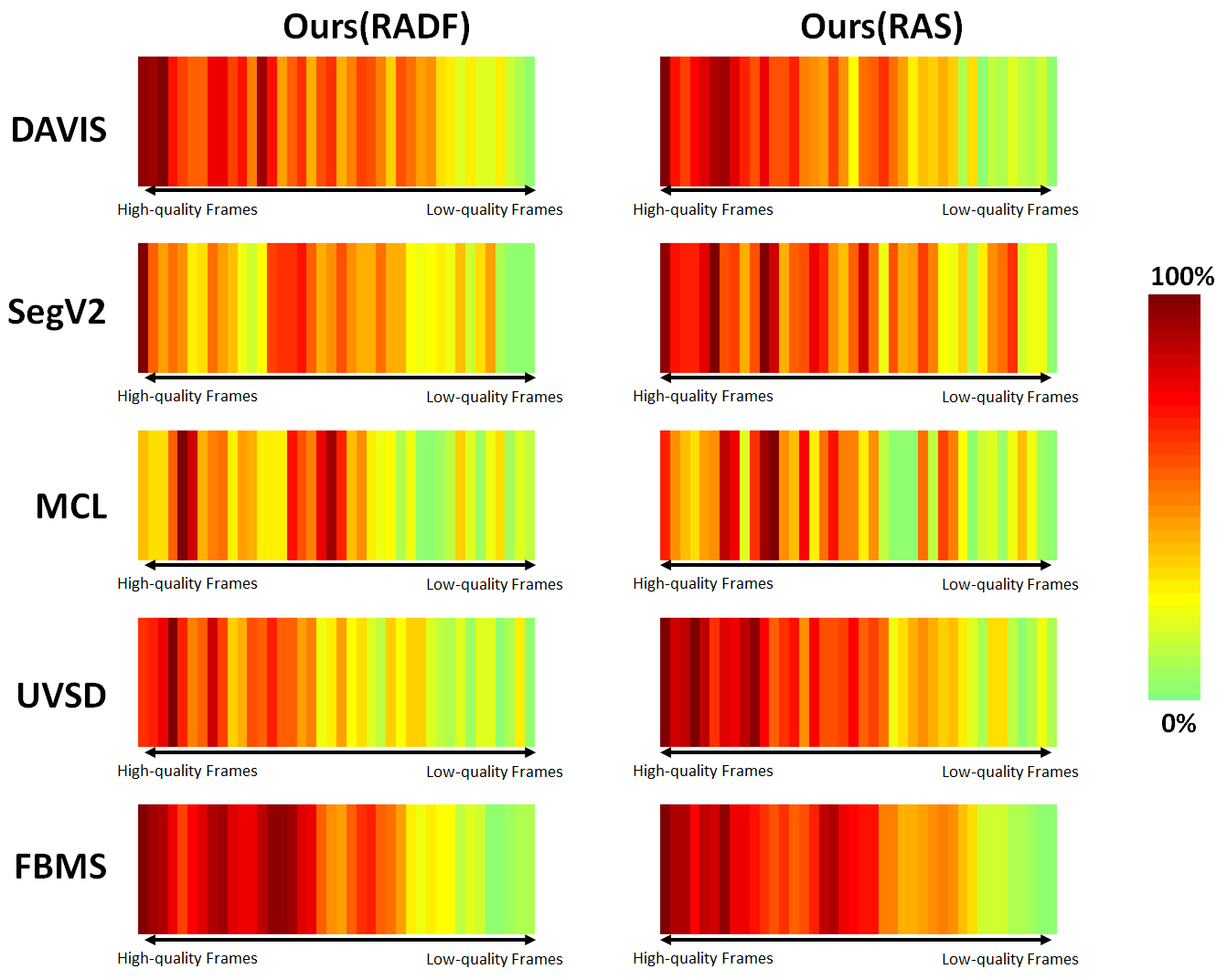}
	\end{center}
	\vspace{-0.4cm}
	\caption{Demonstration of the relationship between the color \& motion saliency consistency and the fused low-level saliency quality, where we have listed the results of two color saliency deep models (RADF~\cite{hu2018recurrent} and RAS~\cite{chen2018eccvRAS}) over five benchmarks.}
	\label{fig:KeyframeDistribution}
	\vspace{0.2cm}
\end{figure}

The key rationale of our weakly supervised online learning is to use the high quality low-level saliency as the pseudo learning ground truth, where we propose a novel keyframe strategy to locate the video frames with high-quality low-level saliency predictions.
Our keyframe strategy is inspired by the phenomenon that only those video frames with both high-quality color and motion saliency may correlate to the cases having high-quality fused low-level saliency.
Thus, we may assume that the degree of consistency between the color and motion saliency maps has a positive relationship with the quality degree of the fused low-level saliency, which motivates us to use such consistency degree to locate those video frames with high-quality low-level saliency predictions.

To further validate this assumption, we demonstrate the correlation between the consistency of color and motion saliency, and the fused low-level saliency quality via multiple quantitative experiments as Fig.~\ref{fig:KeyframeDistribution}.
In these experiments, all video frames are re-ordered according to the qualities of their fused low-level saliency maps, and these qualities are estimated by computing the structural similarity~\cite{SmeasureICCV} between each low-level saliency map and its saliency GT.
Thus, the left side of each sub figure represents those video frames with high-quality fused low-level saliency maps, while the right side represents the low-quality cases.
It should be noted that all video sequences are individually measured, normalized, and aligned to an identical form, containing 40 intervals in total.
\begin{figure*}[!t]
	\begin{center}
		\includegraphics[width=1.0\linewidth]{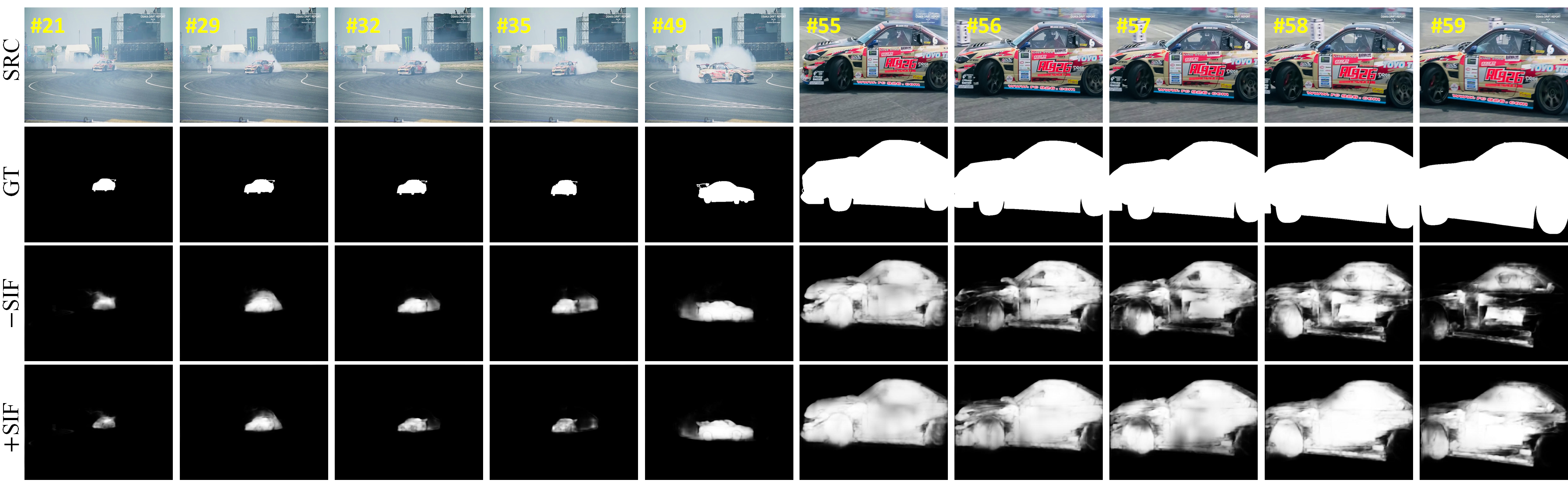}
	\end{center}
	\vspace{-0.4cm}
	\caption{Demonstration of the effectiveness toward the size invariant factor (SIF, Eq.~\ref{eq:overlappingratio}), where GT denotes the ground truth, and ``-SIF'' and ``+SIF'' denote the ``without'' the size invariant factor and ``with'' the size invariant factor respectively.}
	\label{fig:keyframeResultDemo}
	\vspace{-0.4cm}
\end{figure*}

We obtain the overall results by averaging all video sequences of each dataset, where the color of each interval from warm to cold represents the probability to be selected as keyframe.
As shown in Fig.~\ref{fig:KeyframeDistribution}, we have demonstrated the quantitative results of our methods using two different baseline models (i.e. RADF and RAS) over five benchmark datasets, and all results have clearly suggested a positive relationship between the color and motion saliency consistency degree, and the fused low-level saliency quality, which also demonstrates the effectiveness of the proposed keyframe selection strategy.

\begin{figure}[!h]
	\begin{center}
		\includegraphics[width=0.95\linewidth]{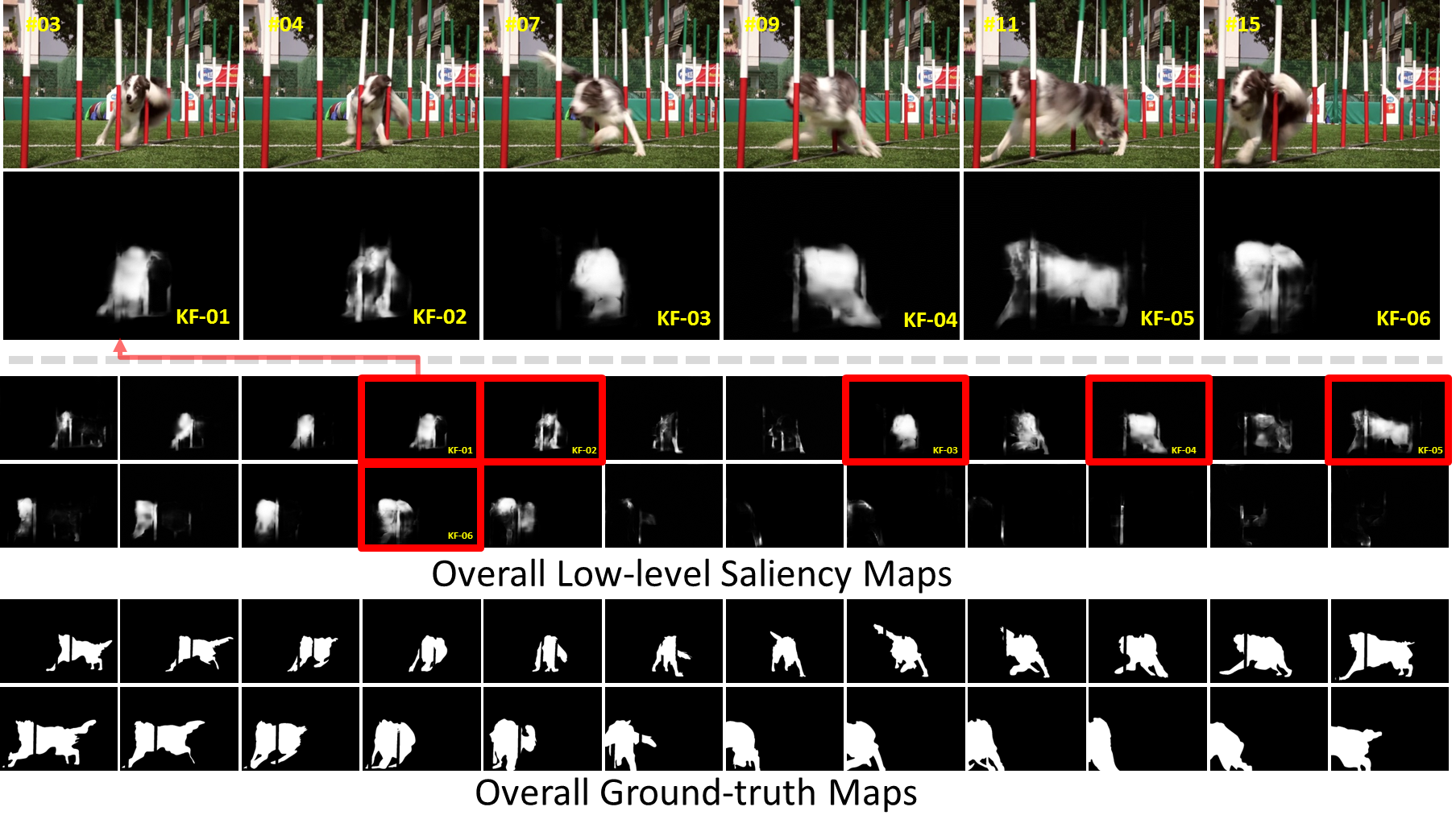}
	\end{center}
	\vspace{-0.6cm}
	\caption{Keyframes selection with low-level saliency maps on the “dog-agility” sequence.}
	\label{fig:KeyframeSample1}
	\vspace{-0.4cm}
\end{figure}

Here we use the non-overlapping ratio $NR$ to measure the degree of consistency between the motion saliency $\hat{\textbf{MS}}$ and the color saliency $\hat{\textbf{CS}}$.
Thus, for the $i$-th video frame, we formulate its non-overlapping ratio $NR$ between the color and motion saliency predictions as Eq.~\ref{eq:overlappingratio} where the value of $NR\in[0,1]$ is inversely related to its degree of quality.
\begin{equation}
\label{eq:overlappingratio}
\begin{split}
&NR_i  = \frac{1}{\underset{size\ invariant\ factor}{\underbrace{||T(\hat{\textbf{MS}}_i + \hat{\textbf{CS}}_i)||_0}}}\\
&\cdot\Bigg |\Bigg |{abs\Big(T(\hat{\textbf{MS}}_i) - T(\hat{\textbf{CS}}_i)\Big)}\odot\frac{1}{T(\hat{\textbf{MS}}_i) + T(\hat{\textbf{CS}}_i)+ {\rm C}}\Bigg |\Bigg |_1,
\end{split}
\end{equation}
where $\rm C$ is a pre-defined constant value (0.001) to avoid division by zero, $T(\cdot)$ represents the hard-threshold filter that assigns 0 to the elements with the value smaller than 0.1, $abs(\cdot)$ denotes the absolute function, and $||\cdot||_0$ and $||\cdot||_1$ represent the L0 and L1 norms, respectively; the ``scale invariant factor'' is used to ensure that our quality assessment is insensitive to the salient region size, and we show its impact over the weakly supervised learning in Fig.~\ref{fig:keyframeResultDemo}. Once the non-overlapping ratio $NR$ is obtained, we select the video frames with $NR<0.6$ as the keyframes, see quantitative proofs in Table~\ref{tab:NewDataNR}.
Moreover, we show the results of keyframe selection strategy of a challenging video sequence in Fig.~\ref{fig:KeyframeSample1}.

\subsection{Self-paced Online Learning}
\label{sec:Self-paced Online Learning}
Given an input video sequence, we use the low-level saliency predictions of the previously located keyframes as the learning pseudo ground truth to weakly fine-tune the color sub-branch.
For the selected keyframes, the high-quality low-level saliency predictions provide a good representation of the long-term spatiotemporal information of the given video sequence.
Furthermore, since the low-level saliency map $\textbf{LS}\in[0,1]$, our self-paced online learning uses the Euclidean loss to replace the conventional cross-entropy loss that is given by Eq.~\ref{eq:Euclideanloss}.
\begin{equation}
\label{eq:Euclideanloss}
L_E(\boldsymbol{\alpha};\boldsymbol{\beta}) = \frac{1}{2N} \sum_{n \in N} \left \| \textbf{LS}_{n} -  \hat{\textbf{EM}}(\boldsymbol{\alpha};\boldsymbol{\beta};\textbf{I}_n) \right \|_2^{2},
\end{equation}
where $L_E$ is the Euclidean loss, $N$ is the training batch size, and $\hat{\textbf{EM}}$ is the estimated saliency map of the color sub-branch.

\begin{figure*}[!t]
	\begin{center}
		\includegraphics[width=0.95\linewidth]{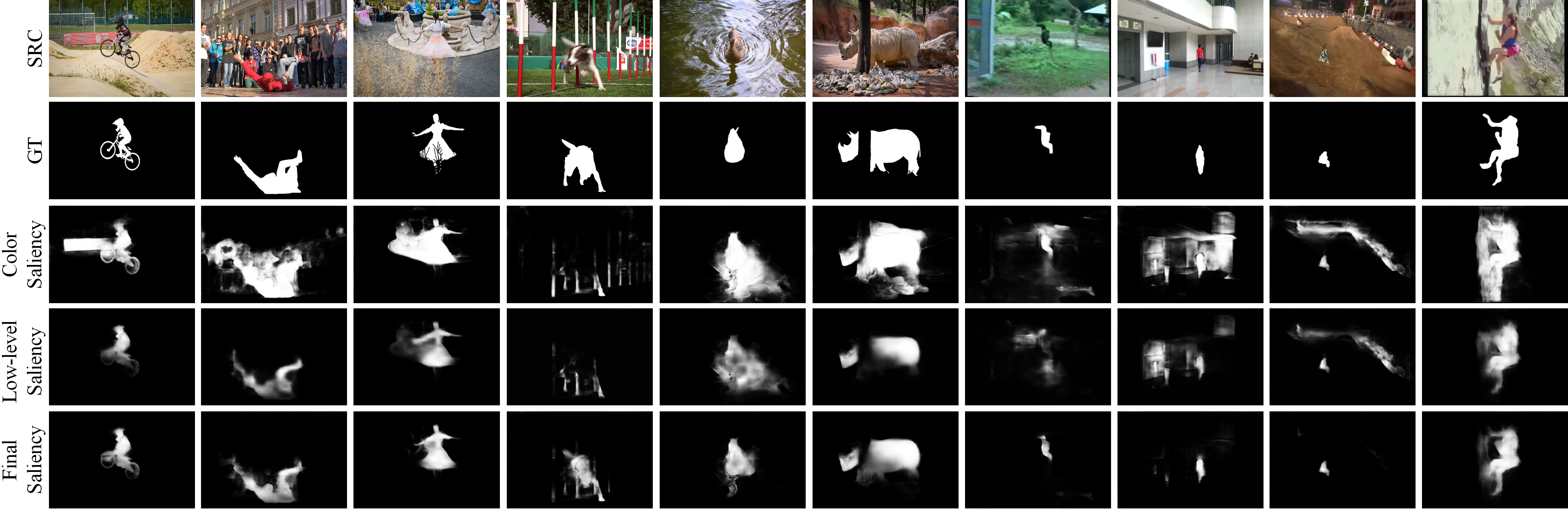}
	\end{center}
	\vspace{-0.4cm}
	\caption{Qualitative illustration of saliency maps obtained via different components, where the 3rd-5th rows respectively denote the color saliency (RADF), the low-level saliency (Eq.~\ref{eq:LowlevelSaliency}) and the final saliency (Eq.~\ref{eq:complementarySaliency}).}
	\label{fig:onlinelearningDemo}
	\vspace{-0.4cm}
\end{figure*}

It should also be noted that we name our weakly supervised learning scheme ``an online manner'' for the following reasons:\\
1) rather than directly conduct model fine-tuning over the original color sub-branch, we fine-tune a ``twin color sub-branch'' with network weights identical to those of the original color sub-branch to achieve improved detection performance;\\
2) our method is data-driven; i.e., the newly fine-tuned color sub-branch is only available for the given video sequence;\\
3) due to a limited problem domain, our fine-tuning procedure for a given video sequence is extremely fast, and can be converged in a very short time.\\
In fact, our self-paced online learning scheme can enhance the deep feature distance between the salient regions and its non-salient nearby surroundings, ensuring its spatial saliency temporally smoothness, as shown in the qualitative demonstrations presented in Fig.~\ref{fig:onlinelearningDemo}.

Once the twin color sub-branch is computed, we attempt to fuse its saliency predictions with the original low-level saliency maps as the final video saliency detection results (Eq.~\ref{eq:complementarySaliency}), in which the fused saliency outperform its inputs slightly, as demonstrated quantitatively in Sec.~\ref{sec:AdaptivenessAnalysis}.
\begin{equation}
\label{eq:complementarySaliency}
\textbf{FS} = \hat{\textbf{EM}^{'}} \odot \textbf{LS},
\end{equation}
where $\textbf{FS}$ denotes our final saliency map, and $\hat{\textbf{EM}^{'}}$ denotes the saliency prediction of our fine-tuned color sub-branch.

\section{Experiments and Evaluations}

\subsection{Datasets}
We evaluate our approach on 5 most widely used public available datasets, including
DAVIS2016(480p)~\cite{Perazzi2016benchmark}, SegV2~\cite{li2013video},
MCL~\cite{Kim:2015},
UVSD~\cite{liu2017saliencySGSP},
and FBMS~\cite{ochs2014segmentation}. All of the ground-truths of these datasets are well-annotated at the pixel level.

%
%
%
%
\begin{table*}[t]
	\centering
	\vspace{-0.3cm}
	\caption{Quantitative comparison results between our method and 15 SOTA methods over 5 public available datasets. The column-wise bests are marked with \textcolor[rgb]{1.000, 0.000, 0.000}{\textbf{red}} color, the 2nd-bests are marked with \textcolor[rgb]{0.000, 0.690, 0.314}{\textbf{green}} color, and the 3rd-bests are marked with \textcolor[rgb]{0.267, 0.447, 0.769}{\textbf{blue}} color.}
	\resizebox{\textwidth}{!}{
		\begin{tabular}{c|c|cccccc|c|c|c|c|c|c|c|c|c|c|c|c|c|c|c}
    \toprule[1pt]
			\multirow{3}[4]{*}{DataSet} & \multirow{3}[4]{*}{Metric} & \multicolumn{6}{c|}{Ours} &  \multicolumn{4}{c|}{2019}   & \multicolumn{4}{c|}{2018}                     & \multicolumn{3}{c|}{2016-2017} & \multicolumn{4}{c}{2015} \\

			\cmidrule{3-23}          &  & \multirow{1}[4]{*}{RADF} &
			\multirow{1}[4]{*}{ResDSS} & \multirow{1}[4]{*}{RAS} & \multirow{1}[4]{*}{CPD} & \multirow{1}[4]{*}{PoolNet} & \multirow{1}[4]{*}{SSAV} & \multicolumn{1}{c}{SSAV} & \multicolumn{1}{c}{CPD} & \multicolumn{1}{c}{ResDSS} & PoolNet & \multicolumn{1}{c}{FL} & \multicolumn{1}{c}{DLVSD} & \multicolumn{1}{c}{RAS} &
			RADF  & \multicolumn{1}{c}{FD} & \multicolumn{1}{c}{SGSP} & RFCN  & \multicolumn{1}{c}{MDF} & \multicolumn{1}{c}{GF} & \multicolumn{1}{c}{MC} & \multicolumn{1}{c}{SA}
			\\

			&       & & & & & & &
			\multicolumn{1}{c}{\cite{fanCVPR2019shifting}} &\multicolumn{1}{c}{\cite{wuCVPR2019cascaded}} & \multicolumn{1}{c}{\cite{CMMTPAMI2019resDSS}} &
			\multicolumn{1}{c|}{\cite{liuCVPR2019simple}} & \multicolumn{1}{c}{\cite{CC_TMM_2018}} & \multicolumn{1}{c}{\cite{wang2018video}} & \multicolumn{1}{c}{\cite{chen2018eccvRAS}} &  \multicolumn{1}{c|}{\cite{hu2018recurrent}} & \multicolumn{1}{c}{\cite{TIP17chen2017video}} &  \multicolumn{1}{c}{\cite{liu2017saliencySGSP}} & \multicolumn{1}{c|}{\cite{wang2016saliency}}  & \multicolumn{1}{c}{\cite{TIP_li2016visual}} & \multicolumn{1}{c}{\cite{TIP15:2015}} & \multicolumn{1}{c}{\cite{Kim:2015}} & \multicolumn{1}{c}{\cite{wang2015saliency}}
			\\
			
			\midrule
			\multirow{3}[2]{*}{DAVIS~\cite{Perazzi2016benchmark}} & \multicolumn{1}{c}{maxF} &  \multicolumn{1}{c}{.885}  & \multicolumn{1}{c}{.874} & \multicolumn{1}{c}{.881} & \multicolumn{1}{c}{\textcolor[rgb]{0.000, 0.439, 0.753}{.896}} & \multicolumn{1}{c}{\textcolor[rgb]{0.000, 0.690, 0.314}{.908}} & \textcolor[rgb]{1.000, 0.000, 0.000}{.912}
			& \multicolumn{1}{c}{.871} & \multicolumn{1}{c}{.827} & \multicolumn{1}{c}{.796} &
			\multicolumn{1}{c}{.826} & \multicolumn{1}{c}{.739} & \multicolumn{1}{c}{.748} & \multicolumn{1}{c}{.780} & \multicolumn{1}{c}{.781} & \multicolumn{1}{c}{.758} & \multicolumn{1}{c}{.707} & \multicolumn{1}{c}{.380} & \multicolumn{1}{c}{.698} & \multicolumn{1}{c}{.621} & \multicolumn{1}{c}{.263} & \multicolumn{1}{c}{.554}
			\\

			& \multicolumn{1}{c}{avgF} & \multicolumn{1}{c}{.815} & \multicolumn{1}{c}{.795} & \multicolumn{1}{c}{.802} & \multicolumn{1}{c}{\textcolor[rgb]{0.000, 0.439, 0.753}{.824}} & \multicolumn{1}{c}{\textcolor[rgb]{0.000, 0.690, 0.314}{.824}} & \textcolor[rgb]{1.000, 0.000, 0.000}{.830}
			& \multicolumn{1}{c}{.823} &\multicolumn{1}{c}{.794} & \multicolumn{1}{c}{.731} &
			\multicolumn{1}{c}{.794} &  \multicolumn{1}{c}{.642} & \multicolumn{1}{c}{.669} & \multicolumn{1}{c}{.724} &  \multicolumn{1}{c}{.701} & \multicolumn{1}{c}{.696} & \multicolumn{1}{c}{.522} & \multicolumn{1}{c}{.363} & \multicolumn{1}{c}{.662} & \multicolumn{1}{c}{.517} & \multicolumn{1}{c}{.177} & \multicolumn{1}{c}{.469}
			\\

			& \multicolumn{1}{c}{MAE} & \multicolumn{1}{c}{.029} & \multicolumn{1}{c}{.033} & \multicolumn{1}{c}{.034} & \multicolumn{1}{c}{\textcolor[rgb]{0.000, 0.690, 0.314}{.028}} &
			\multicolumn{1}{c}{.029} & \textcolor[rgb]{1.000, 0.000, 0.000}{.028}
			& \multicolumn{1}{c}{\textcolor[rgb]{0.000, 0.690, 0.314}{.028}} & \multicolumn{1}{c}{.033} & \multicolumn{1}{c}{.045} & \multicolumn{1}{c}{.039} &  \multicolumn{1}{c}{.074} & \multicolumn{1}{c}{.059} & \multicolumn{1}{c}{.048} & \multicolumn{1}{c}{.055} & \multicolumn{1}{c}{.055} & \multicolumn{1}{c}{.136} & \multicolumn{1}{c}{.082} & \multicolumn{1}{c}{.073} & \multicolumn{1}{c}{.099} & \multicolumn{1}{c}{.244} & \multicolumn{1}{c}{.101}
			\\

			\midrule
			\multirow{3}[2]{*}{SegV2~\cite{li2013video}} & \multicolumn{1}{c}{maxF} & \multicolumn{1}{c}{.842} & \multicolumn{1}{c}{.853} & \multicolumn{1}{c}{.825} & \multicolumn{1}{c}{\textcolor[rgb]{1, 0, 0}{.884}}
			& \multicolumn{1}{c}{\textcolor[rgb]{0.000, 0.690, 0.314}{.879}} & \textcolor[rgb]{0.000, 0.439, 0.753}{.878}
			& \multicolumn{1}{c}{.813} &\multicolumn{1}{c}{.828} & \multicolumn{1}{c}{.836} &  \multicolumn{1}{c}{.785} &  \multicolumn{1}{c}{.842} & \multicolumn{1}{c}{.747} & \multicolumn{1}{c}{.761} &  \multicolumn{1}{c}{.807} & \multicolumn{1}{c}{.820} &  \multicolumn{1}{c}{.691} & \multicolumn{1}{c}{.368} & \multicolumn{1}{c}{.683} & \multicolumn{1}{c}{.739} & \multicolumn{1}{c}{.500} & \multicolumn{1}{c}{.716}
			\\

			& \multicolumn{1}{c}{avgF} & \multicolumn{1}{c}{.765} & \multicolumn{1}{c}{.762} & \multicolumn{1}{c}{.730} & \multicolumn{1}{c}{\textcolor[rgb]{0.000, 0.690, 0.314}{.779}} & \multicolumn{1}{c}{\textcolor[rgb]{0.000, 0.439, 0.753}{.775}} &
			.764
			& \multicolumn{1}{c}{.752}
			& \multicolumn{1}{c}{\textcolor[rgb]{1, 0, 0}{.796}} & \multicolumn{1}{c}{.755} & \multicolumn{1}{c}{.745}  &  \multicolumn{1}{c}{.741} & \multicolumn{1}{c}{.627} & \multicolumn{1}{c}{.705} & \multicolumn{1}{c}{.724} & \multicolumn{1}{c}{.754} & \multicolumn{1}{c}{.505} & \multicolumn{1}{c}{.313} & \multicolumn{1}{c}{.648} & \multicolumn{1}{c}{.603} & \multicolumn{1}{c}{.304} & \multicolumn{1}{c}{.557}
			\\

			& \multicolumn{1}{c}{MAE} &  \multicolumn{1}{c}{.026} & \multicolumn{1}{c}{.027} & \multicolumn{1}{c}{.028} & \multicolumn{1}{c}{\textcolor[rgb]{0.000, 0.439, 0.753}{.025}} &
			\multicolumn{1}{c}{.026} & .026
			& \multicolumn{1}{c}{.026}
			& \multicolumn{1}{c}{\textcolor[rgb]{1.000, 0.000, 0.000}{.021}} & \multicolumn{1}{c}{.031} &\multicolumn{1}{c}{\textcolor[rgb]{0.000, 0.690, 0.314}{.022}} &  \multicolumn{1}{c}{.042} & \multicolumn{1}{c}{.044} & \multicolumn{1}{c}{.031} & \multicolumn{1}{c}{.034} & \multicolumn{1}{c}{.033} &  \multicolumn{1}{c}{.116} & \multicolumn{1}{c}{.055} & \multicolumn{1}{c}{.053} & \multicolumn{1}{c}{.081} & \multicolumn{1}{c}{.163} & \multicolumn{1}{c}{.086}
			\\

			\midrule
			\multirow{3}[2]{*}{MCL~\cite{Kim:2015}} & \multicolumn{1}{c}{maxF} &  \multicolumn{1}{c}{\textcolor[rgb]{0.000, 0.439, 0.753}{.793}} & \multicolumn{1}{c}{.749} & \multicolumn{1}{c}{.791} & \multicolumn{1}{c}{.790} & \multicolumn{1}{c}{\textcolor[rgb]{0.000, 0.690, 0.314}{.812}} &
			\textcolor[rgb]{1, 0, 0}{.846}
			& \multicolumn{1}{c}{.745} & \multicolumn{1}{c}{.656} & \multicolumn{1}{c}{.628} &\multicolumn{1}{c}{.644} &  \multicolumn{1}{c}{.727} & \multicolumn{1}{c}{.600} & \multicolumn{1}{c}{.670} &  \multicolumn{1}{c}{.611} & \multicolumn{1}{c}{.707} &  \multicolumn{1}{c}{.682} & \multicolumn{1}{c}{.203} & \multicolumn{1}{c}{.601} & \multicolumn{1}{c}{.454} & \multicolumn{1}{c}{.483} & \multicolumn{1}{c}{.473}
			\\

			& \multicolumn{1}{c}{avgF} &\multicolumn{1}{c}{.679} & \multicolumn{1}{c}{.609} & \multicolumn{1}{c}{.652} & \multicolumn{1}{c}{.672} & \multicolumn{1}{c}{\textcolor[rgb]{0.000, 0.439, 0.753}{.683}}
			& \textcolor[rgb]{0.000, 0.690, 0.314}{.693}
			& \multicolumn{1}{c}{\textcolor[rgb]{1, 0, 0}{.708}} &\multicolumn{1}{c}{.629}  & \multicolumn{1}{c}{.564} & \multicolumn{1}{c}{.618}&  \multicolumn{1}{c}{.664} & \multicolumn{1}{c}{.499} & \multicolumn{1}{c}{.610}  & \multicolumn{1}{c}{.555} & \multicolumn{1}{c}{.634} &  \multicolumn{1}{c}{.521} & \multicolumn{1}{c}{.173} & \multicolumn{1}{c}{.574} & \multicolumn{1}{c}{.390} & \multicolumn{1}{c}{.289} & \multicolumn{1}{c}{.366}
			\\

			& \multicolumn{1}{c}{MAE} &\multicolumn{1}{c}{.041} & \multicolumn{1}{c}{.041} & \multicolumn{1}{c}{.039} & \multicolumn{1}{c}{\textcolor[rgb]{0.000, 0.439, 0.753}{.038}} &
			\multicolumn{1}{c}{.039} & \textcolor[rgb]{0.000, 0.690, 0.314}{.034}
			& \multicolumn{1}{c}{\textcolor[rgb]{1, 0, 0}{.028}} & \multicolumn{1}{c}{.043} & \multicolumn{1}{c}{.044} &
			\multicolumn{1}{c}{.051} &  \multicolumn{1}{c}{.049} & \multicolumn{1}{c}{.056} & \multicolumn{1}{c}{.048} &  \multicolumn{1}{c}{.071} & \multicolumn{1}{c}{.053} & \multicolumn{1}{c}{.092} & \multicolumn{1}{c}{.067} & \multicolumn{1}{c}{.045} & \multicolumn{1}{c}{.136} & \multicolumn{1}{c}{.167} & \multicolumn{1}{c}{.139}
			\\

			\midrule
			\multirow{3}[2]{*}{UVSD~\cite{liu2017saliencySGSP}} & \multicolumn{1}{c}{maxF} & \multicolumn{1}{c}{.703}& \multicolumn{1}{c}{.694} & \multicolumn{1}{c}{.712} & \multicolumn{1}{c}{.713} & \multicolumn{1}{c}{\textcolor[rgb]{0.000, 0.439, 0.753}{.733}} &
			\textcolor[rgb]{0.000, 0.690, 0.314}{.762}
			& \multicolumn{1}{c}{\textcolor[rgb]{1.000, 0.000, 0.000}{.811}} & \multicolumn{1}{c}{.674} & \multicolumn{1}{c}{.616} & \multicolumn{1}{c}{.615} &  \multicolumn{1}{c}{.614} & \multicolumn{1}{c}{.586} & \multicolumn{1}{c}{.665} &  \multicolumn{1}{c}{.545} & \multicolumn{1}{c}{.611} &  \multicolumn{1}{c}{.615} & \multicolumn{1}{c}{.202} & \multicolumn{1}{c}{.523} & \multicolumn{1}{c}{.502} & \multicolumn{1}{c}{.300} & \multicolumn{1}{c}{.485}
			\\

			& \multicolumn{1}{c}{avgF} & \multicolumn{1}{c}{.608} & \multicolumn{1}{c}{.592} & \multicolumn{1}{c}{.609} & \multicolumn{1}{c}{.625} &
			\multicolumn{1}{c}{.614} &
			\textcolor[rgb]{0.000, 0.690, 0.314}{.643}
			& \multicolumn{1}{c}{\textcolor[rgb]{1, 0, 0}{.736}} & \multicolumn{1}{c}{\textcolor[rgb]{0.000, 0.439, 0.753}{.643}} & \multicolumn{1}{c}{.559} & \multicolumn{1}{c}{.577} &  \multicolumn{1}{c}{.564} & \multicolumn{1}{c}{.497} & \multicolumn{1}{c}{.610} & \multicolumn{1}{c}{.484} & \multicolumn{1}{c}{.559} & \multicolumn{1}{c}{.427} & \multicolumn{1}{c}{.176} & \multicolumn{1}{c}{.503} & \multicolumn{1}{c}{.420} & \multicolumn{1}{c}{.187} & \multicolumn{1}{c}{.396}
			\\

			& \multicolumn{1}{c}{MAE} &  \multicolumn{1}{c}{.038} & \multicolumn{1}{c}{.037} & \multicolumn{1}{c}{.037} & \multicolumn{1}{c}{\textcolor[rgb]{0.000, 0.439, 0.753}{.031}} & \multicolumn{1}{c}{.035} & \textcolor[rgb]{0.000, 0.690, 0.314}{.029}
			& \multicolumn{1}{c}{\textcolor[rgb]{1.000, 0.000, 0.000}{.023}}
			& \multicolumn{1}{c}{.038} & \multicolumn{1}{c}{.047} & \multicolumn{1}{c}{.041} &  \multicolumn{1}{c}{.070} & \multicolumn{1}{c}{.056} & \multicolumn{1}{c}{.043} &  \multicolumn{1}{c}{.074} & \multicolumn{1}{c}{.054} &  \multicolumn{1}{c}{.156} & \multicolumn{1}{c}{.065} & \multicolumn{1}{c}{.059} & \multicolumn{1}{c}{.131} & \multicolumn{1}{c}{.173} & \multicolumn{1}{c}{.105}
			\\

			\midrule
			\multirow{3}[2]{*}{FBMS~\cite{ochs2014segmentation}} & \multicolumn{1}{c}{maxF} & \multicolumn{1}{c}{.824} & \multicolumn{1}{c}{.800} & \multicolumn{1}{c}{.811} & \multicolumn{1}{c}{.839} & \multicolumn{1}{c}{\textcolor[rgb]{0.000, 0.690, 0.314}{.859}} &
			.858
			& \multicolumn{1}{c}{\textcolor[rgb]{1, 0, 0}{.869}}
			& \multicolumn{1}{c}{.833} &\multicolumn{1}{c}{.790} & \multicolumn{1}{c}{\textcolor[rgb]{0.000, 0.439, 0.753}{.858}} &  \multicolumn{1}{c}{.676} & \multicolumn{1}{c}{.762} & \multicolumn{1}{c}{.801} & \multicolumn{1}{c}{.776} & \multicolumn{1}{c}{.692} & \multicolumn{1}{c}{.671} & \multicolumn{1}{c}{.422} & \multicolumn{1}{c}{.713} & \multicolumn{1}{c}{.602} & \multicolumn{1}{c}{.363} & \multicolumn{1}{c}{.569}
			\\

			& \multicolumn{1}{c}{avgF} & \multicolumn{1}{c}{.709} & \multicolumn{1}{c}{.682} & \multicolumn{1}{c}{.693}
			& \multicolumn{1}{c}{.718} &
			\multicolumn{1}{c}{.732} &
			.728
			& \multicolumn{1}{c}{\textcolor[rgb]{0.000, 0.690, 0.314}{.832}}
			& \multicolumn{1}{c}{\textcolor[rgb]{0.000, 0.439, 0.753}{.809}} &  \multicolumn{1}{c}{.749} & \multicolumn{1}{c}{\textcolor[rgb]{1.000, 0.000, 0.000}{.835}} &  \multicolumn{1}{c}{.615} & \multicolumn{1}{c}{.696} & \multicolumn{1}{c}{.757}  & \multicolumn{1}{c}{.740} & \multicolumn{1}{c}{.649} & \multicolumn{1}{c}{.527} & \multicolumn{1}{c}{.403} & \multicolumn{1}{c}{.653} & \multicolumn{1}{c}{.497} & \multicolumn{1}{c}{.224} & \multicolumn{1}{c}{.473}
			\\

			& \multicolumn{1}{c}{MAE} & \multicolumn{1}{c}{.101} & \multicolumn{1}{c}{.106} & \multicolumn{1}{c}{.104}
			& \multicolumn{1}{c}{.092} & \multicolumn{1}{c}{.089}
			& .090 & \multicolumn{1}{c}{\textcolor[rgb]{0.000, 0.690, 0.314}{.045}}
			& \multicolumn{1}{c}{\textcolor[rgb]{0.000, 0.439, 0.753}{.057}} & \multicolumn{1}{c}{.087} &
			\multicolumn{1}{c}{\textcolor[rgb]{1.000, 0.000, 0.000}{.044}} &  \multicolumn{1}{c}{.163} & \multicolumn{1}{c}{.105} & \multicolumn{1}{c}{.086}  & \multicolumn{1}{c}{.095} & \multicolumn{1}{c}{.132} & \multicolumn{1}{c}{.181} & \multicolumn{1}{c}{.154} & \multicolumn{1}{c}{.118} & \multicolumn{1}{c}{.177} & \multicolumn{1}{c}{.229} & \multicolumn{1}{c}{.185}
			\\

    \toprule[1pt]
		\end{tabular}%
	}
	\label{tab:NewData}%
\end{table*}%

\begin{table*}[htbp]
	\definecolor{mygray}{gray}{.9}
	\vspace{-0.3cm}
	\caption{Quantitative component evaluations toward our 6 re-trained saliency deep models (5 image saliency models and 1 video saliency model) including RADF, ResDSS, RAS, CPD, PoolNet, SSAV over 5 datasets. }
	\resizebox{\textwidth}{!}{
		\begin{tabular}{c|c|c|c|c|c|c|c|c|c|c|c|c|c|c|c|c|c|c|c}
			
    \toprule[1pt]
			\multirow{2}{*}{DataSet} & \multirow{2}{*}{Metric}
			& \multicolumn{1}{c}{Ours} & \multicolumn{1}{c}{Lowlevel} & Original &
			\multicolumn{1}{c}{Ours} & \multicolumn{1}{c}{Lowlevel} & Original  & \multicolumn{1}{c}{Ours} & \multicolumn{1}{c}{Lowlevel} & Original  &
			\multicolumn{1}{c}{Ours}  & \multicolumn{1}{c}{Lowlevel} & Original &
			\multicolumn{1}{c}{Ours}  & \multicolumn{1}{c}{Lowlevel} & Original &
			\multicolumn{1}{c}{Ours}  & \multicolumn{1}{c}{Lowlevel} & Original
			\\

			&    & \multicolumn{1}{c}{RADF} & \multicolumn{1}{c}{Saliency} & RADF & \multicolumn{1}{c}{ResDSS} & \multicolumn{1}{c}{Saliency} & ResDSS &
			\multicolumn{1}{c}{RAS} & \multicolumn{1}{c}{Saliency} & RAS &
			\multicolumn{1}{c}{CPD} & \multicolumn{1}{c}{Saliency} & CPD &
			\multicolumn{1}{c}{PoolNet} & \multicolumn{1}{c}{Saliency} & PoolNet &
			\multicolumn{1}{c}{SSAV} & \multicolumn{1}{c}{Saliency} & SSAV
			\\
			
			\midrule
			\multirow{2}[3]{*}{DAVIS~\cite{Perazzi2016benchmark}} & \multicolumn{1}{c|}{maxF} &
			\multicolumn{1}{c}{.885}  & \multicolumn{1}{c}{.868} & .781& \multicolumn{1}{c}{.874} & \multicolumn{1}{c}{.864} & .796 &
			\multicolumn{1}{c}{.881} & \multicolumn{1}{c}{.870} & .780 &
			\multicolumn{1}{c}{.896} & \multicolumn{1}{c}{.878} & .827 &
			
			\multicolumn{1}{c}{.908} & \multicolumn{1}{c}{.907} & .826 &
			\multicolumn{1}{c}{.912} & \multicolumn{1}{c}{.906} & .871
			\\
			
			& \multicolumn{1}{c|}{AvgF} &
			\multicolumn{1}{c}{\cellcolor{mygray}.815} & \multicolumn{1}{c}{\cellcolor{mygray}.781} & \multicolumn{1}{c|}{\cellcolor{mygray}.701} & \multicolumn{1}{c}{\cellcolor{mygray}.795} & \multicolumn{1}{c}{\cellcolor{mygray}.769} & \multicolumn{1}{c|}{\cellcolor{mygray}.731} &
			\multicolumn{1}{c}{\cellcolor{mygray}.802} & \multicolumn{1}{c}{\cellcolor{mygray}.776} &
			\multicolumn{1}{c|}{\cellcolor{mygray}.724} &
			\multicolumn{1}{c}{\cellcolor{mygray}.824} & \multicolumn{1}{c}{\cellcolor{mygray}.793} & \multicolumn{1}{c|}{\cellcolor{mygray}.794}&
			
			\multicolumn{1}{c}{\cellcolor{mygray}.824} & \multicolumn{1}{c}{\cellcolor{mygray}.807} & \multicolumn{1}{c|}{\cellcolor{mygray}.794}&
			\multicolumn{1}{c}{\cellcolor{mygray}.830} & \multicolumn{1}{c}{\cellcolor{mygray}.811} & \multicolumn{1}{c}{\cellcolor{mygray}.823}
			\\

			&\multicolumn{1}{c|}{MAE} &
			\multicolumn{1}{c}{.029}  & \multicolumn{1}{c}{.037} & .055& \multicolumn{1}{c}{.033} & \multicolumn{1}{c}{.036} & .045 &
			\multicolumn{1}{c}{.034} & \multicolumn{1}{c}{.037} & .048 &
			\multicolumn{1}{c}{.028} & \multicolumn{1}{c}{.034} & .033 &
			
			\multicolumn{1}{c}{.029} & \multicolumn{1}{c}{.045} & .039 &
			\multicolumn{1}{c}{.028} & \multicolumn{1}{c}{.031} & .028
			\\

			\midrule
			\multirow{2}[3]{*}{SegV2~\cite{li2013video}} & \multicolumn{1}{c|}{maxF} &
			\multicolumn{1}{c}{.842}  & \multicolumn{1}{c}{.827} & \multicolumn{1}{c|}{.807}& \multicolumn{1}{c}{.853} & \multicolumn{1}{c}{.835} & \multicolumn{1}{c|}{.836} &
			\multicolumn{1}{c}{.825} & \multicolumn{1}{c}{.790} & \multicolumn{1}{c|}{.761} &
			\multicolumn{1}{c}{.884} & \multicolumn{1}{c}{.857} & \multicolumn{1}{c|}{.828}&
			
			\multicolumn{1}{c}{.879} & \multicolumn{1}{c}{.849} & \multicolumn{1}{c|}{.785}&
			\multicolumn{1}{c}{.878} & \multicolumn{1}{c}{.848} & \multicolumn{1}{c}{.813}
			\\
			
			& \multicolumn{1}{c|}{AvgF} &
			\multicolumn{1}{c}{\cellcolor{mygray}.765} & \multicolumn{1}{c}{\cellcolor{mygray}.718} & \multicolumn{1}{c|}{\cellcolor{mygray}.724} & \multicolumn{1}{c}{\cellcolor{mygray}.762} & \multicolumn{1}{c}{\cellcolor{mygray}.716} & \multicolumn{1}{c|}{\cellcolor{mygray}.755} &
			\multicolumn{1}{c}{\cellcolor{mygray}.730} & \multicolumn{1}{c}{\cellcolor{mygray}.666} &
			\multicolumn{1}{c|}{\cellcolor{mygray}.705} &
			\multicolumn{1}{c}{\cellcolor{mygray}.779} & \multicolumn{1}{c}{\cellcolor{mygray}.726} & \multicolumn{1}{c|}{\cellcolor{mygray}.796}&
			
			\multicolumn{1}{c}{\cellcolor{mygray}.775} & \multicolumn{1}{c}{\cellcolor{mygray}.712} & \multicolumn{1}{c|}{\cellcolor{mygray}.745}&
			\multicolumn{1}{c}{\cellcolor{mygray}.764} & \multicolumn{1}{c}{\cellcolor{mygray}.696} & \multicolumn{1}{c}{\cellcolor{mygray}.752}
			\\

			&\multicolumn{1}{c|}{MAE} &
			\multicolumn{1}{c}{.026}  & \multicolumn{1}{c}{.034} & \multicolumn{1}{c|}{.034}& \multicolumn{1}{c}{.027} & \multicolumn{1}{c}{.032} & \multicolumn{1}{c|}{.031} &
			\multicolumn{1}{c}{.028} & \multicolumn{1}{c}{.034} & \multicolumn{1}{c|}{.031} &
			\multicolumn{1}{c}{.025} & \multicolumn{1}{c}{.033} & \multicolumn{1}{c|}{.021}&
			
			\multicolumn{1}{c}{.026} & \multicolumn{1}{c}{.035} & \multicolumn{1}{c|}{.022}&
			\multicolumn{1}{c}{.026} & \multicolumn{1}{c}{.033} & \multicolumn{1}{c}{.026}
			\\

			\midrule
			\multirow{2}[3]{*}{MCL~\cite{Kim:2015}} & \multicolumn{1}{c|}{maxF} &
			\multicolumn{1}{c}{.793}  & \multicolumn{1}{c}{.761} & \multicolumn{1}{c|}{.611}& \multicolumn{1}{c}{.749} & \multicolumn{1}{c}{.737} & \multicolumn{1}{c|}{.628} &
			\multicolumn{1}{c}{.791} & \multicolumn{1}{c}{.757} & \multicolumn{1}{c|}{.670} &
			\multicolumn{1}{c}{.790} & \multicolumn{1}{c}{.746} & \multicolumn{1}{c|}{.656}&
			
			\multicolumn{1}{c}{.812} & \multicolumn{1}{c}{.762} & \multicolumn{1}{c|}{.644}&
			\multicolumn{1}{c}{.846} & \multicolumn{1}{c}{.821} & \multicolumn{1}{c}{.745}
			\\

			& \multicolumn{1}{c|}{AvgF} &
			\multicolumn{1}{c}{\cellcolor{mygray}.679} & \multicolumn{1}{c}{\cellcolor{mygray}.619} & \multicolumn{1}{c|}{\cellcolor{mygray}.555} & \multicolumn{1}{c}{\cellcolor{mygray}.609} & \multicolumn{1}{c}{\cellcolor{mygray}.575} & \multicolumn{1}{c|}{\cellcolor{mygray}.564} &
			\multicolumn{1}{c}{\cellcolor{mygray}.652} & \multicolumn{1}{c}{\cellcolor{mygray}.599} &
			\multicolumn{1}{c|}{\cellcolor{mygray}.610} &
			\multicolumn{1}{c}{\cellcolor{mygray}.672} & \multicolumn{1}{c}{\cellcolor{mygray}.614} & \multicolumn{1}{c|}{\cellcolor{mygray}.629}&
			
			\multicolumn{1}{c}{\cellcolor{mygray}.683} & \multicolumn{1}{c}{\cellcolor{mygray}.627} & \multicolumn{1}{c|}{\cellcolor{mygray}.618}	&
			\multicolumn{1}{c}{\cellcolor{mygray}.693} & \multicolumn{1}{c}{\cellcolor{mygray}.653} & \multicolumn{1}{c}{\cellcolor{mygray}.708}	
			\\

			&\multicolumn{1}{c|}{MAE} &
			\multicolumn{1}{c}{.041}  & \multicolumn{1}{c}{.048} & \multicolumn{1}{c|}{.071}& \multicolumn{1}{c}{.041} & \multicolumn{1}{c}{.044} & \multicolumn{1}{c|}{.044} &
			\multicolumn{1}{c}{.039} & \multicolumn{1}{c}{.046} & \multicolumn{1}{c|}{.048} &
			\multicolumn{1}{c}{.038} & \multicolumn{1}{c}{.045} & \multicolumn{1}{c|}{.043}&
			
			\multicolumn{1}{c}{.039} & \multicolumn{1}{c}{.048} & \multicolumn{1}{c|}{.051}&
			\multicolumn{1}{c}{.034} & \multicolumn{1}{c}{.038} & \multicolumn{1}{c}{.028}
			\\

			\midrule
			\multirow{2}[3]{*}{UVSD~\cite{liu2017saliencySGSP}} & \multicolumn{1}{c|}{maxF} &
			\multicolumn{1}{c}{.703}  & \multicolumn{1}{c}{.666} & \multicolumn{1}{c|}{.545}& \multicolumn{1}{c}{.694} & \multicolumn{1}{c}{.660} & \multicolumn{1}{c|}{.619} &
			\multicolumn{1}{c}{.712} & \multicolumn{1}{c}{.683} & \multicolumn{1}{c|}{.665} &
			\multicolumn{1}{c}{.713} & \multicolumn{1}{c}{.708} & \multicolumn{1}{c|}{.674}&
			
			\multicolumn{1}{c}{.733} & \multicolumn{1}{c}{.697} & \multicolumn{1}{c|}{.615}&
			\multicolumn{1}{c}{.762} & \multicolumn{1}{c}{.790} & \multicolumn{1}{c}{.811}
			\\

			& \multicolumn{1}{c|}{AvgF} &
			\multicolumn{1}{c}{\cellcolor{mygray}.608} & \multicolumn{1}{c}{\cellcolor{mygray}.570} & \multicolumn{1}{c|}{\cellcolor{mygray}.484} & \multicolumn{1}{c}{\cellcolor{mygray}.592} & \multicolumn{1}{c}{\cellcolor{mygray}.554} & \multicolumn{1}{c|}{\cellcolor{mygray}.559} &
			\multicolumn{1}{c}{\cellcolor{mygray}.609} & \multicolumn{1}{c}{\cellcolor{mygray}.581} &
			\multicolumn{1}{c|}{\cellcolor{mygray}.610} &
			\multicolumn{1}{c}{\cellcolor{mygray}.625} & \multicolumn{1}{c}{\cellcolor{mygray}.592} & \multicolumn{1}{c|}{\cellcolor{mygray}.643}&
			
			\multicolumn{1}{c}{\cellcolor{mygray}.614} & \multicolumn{1}{c}{\cellcolor{mygray}.575} & \multicolumn{1}{c|}{\cellcolor{mygray}.577}&
			\multicolumn{1}{c}{\cellcolor{mygray}.643} & \multicolumn{1}{c}{\cellcolor{mygray}.654} & \multicolumn{1}{c}{\cellcolor{mygray}.736}
			\\
			
			&\multicolumn{1}{c|}{MAE} &
			\multicolumn{1}{c}{.038}  & \multicolumn{1}{c}{.041} & \multicolumn{1}{c|}{.074}& \multicolumn{1}{c}{.037} & \multicolumn{1}{c}{.038} & \multicolumn{1}{c|}{.047} &
			\multicolumn{1}{c}{.037} & \multicolumn{1}{c}{.039} & \multicolumn{1}{c|}{.043} &
			\multicolumn{1}{c}{.031} & \multicolumn{1}{c}{.037} & \multicolumn{1}{c|}{.038}&
			
			\multicolumn{1}{c}{.035} & \multicolumn{1}{c}{.040} & \multicolumn{1}{c|}{.041}&
			\multicolumn{1}{c}{.029} & \multicolumn{1}{c}{.031} & \multicolumn{1}{c}{.023}
			\\

			\midrule
			\multirow{2}[3]{*}{FBMS~\cite{ochs2014segmentation}} & \multicolumn{1}{c|}{maxF} &
			\multicolumn{1}{c}{.824}  & \multicolumn{1}{c}{.795} & \multicolumn{1}{c|}{.776}& \multicolumn{1}{c}{.800} & \multicolumn{1}{c}{.780} & \multicolumn{1}{c|}{.790} &
			\multicolumn{1}{c}{.811} & \multicolumn{1}{c}{.797} & \multicolumn{1}{c|}{.801} &
			\multicolumn{1}{c}{.839} & \multicolumn{1}{c}{.819} & \multicolumn{1}{c|}{.833}&
			
			\multicolumn{1}{c}{.859} & \multicolumn{1}{c}{.843} & \multicolumn{1}{c|}{.858}&
			\multicolumn{1}{c}{.858} & \multicolumn{1}{c}{.841} & \multicolumn{1}{c}{.869}
			\\

			& \multicolumn{1}{c|}{AvgF} &
			\multicolumn{1}{c}{\cellcolor{mygray}.709} & \multicolumn{1}{c}{\cellcolor{mygray}.664} & \multicolumn{1}{c|}{\cellcolor{mygray}.740} & \multicolumn{1}{c}{\cellcolor{mygray}.682} & \multicolumn{1}{c}{\cellcolor{mygray}.648} & \multicolumn{1}{c|}{\cellcolor{mygray}.749} &
			\multicolumn{1}{c}{\cellcolor{mygray}.693} & \multicolumn{1}{c}{\cellcolor{mygray}.661} &
			\multicolumn{1}{c|}{\cellcolor{mygray}.757} &
			\multicolumn{1}{c}{\cellcolor{mygray}.718} & \multicolumn{1}{c}{\cellcolor{mygray}.691} & \multicolumn{1}{c|}{\cellcolor{mygray}.809}&
			
			\multicolumn{1}{c}{\cellcolor{mygray}.732} & \multicolumn{1}{c}{\cellcolor{mygray}.705} & \multicolumn{1}{c|}{\cellcolor{mygray}.835}&
			\multicolumn{1}{c}{\cellcolor{mygray}.728} & \multicolumn{1}{c}{\cellcolor{mygray}.698} & \multicolumn{1}{c}{\cellcolor{mygray}.832}
			\\
			
			&\multicolumn{1}{c|}{MAE} &
			\multicolumn{1}{c}{.101}  & \multicolumn{1}{c}{.113} & \multicolumn{1}{c|}{.095}& \multicolumn{1}{c}{.106} & \multicolumn{1}{c}{.112} & \multicolumn{1}{c|}{.087} &
			\multicolumn{1}{c}{.104} & \multicolumn{1}{c}{.112} & \multicolumn{1}{c|}{.087} &
			\multicolumn{1}{c}{.092} & \multicolumn{1}{c}{.105} & \multicolumn{1}{c|}{.057}&
			
			\multicolumn{1}{c}{.089} & \multicolumn{1}{c}{.107} & \multicolumn{1}{c|}{.044}&
			\multicolumn{1}{c}{.090} & \multicolumn{1}{c}{.102} & \multicolumn{1}{c}{.045}
			\\

    \toprule[1pt]
		\end{tabular}%
	}
    \vspace{-0.4cm}
	\label{tab:NewDataComponent}%
\end{table*}%

\subsection{Evaluation Metrics}
To better verify and validate the performance of our method, we use
3 widely adopted metrics, namely, the mean absolute error (MAE), the maximum F-measure value (maxF) and the average F-measure value (avgF).

We segment the video
saliency detection results of different methods using the same
integer threshold (${T}\in[0,255]$). Then, the regions are labeled as 1 when their saliency values are greater than ${T}$ and the other regions are set to 0.
Since the recall rate
is inversely proportional to the precision, the tendency of the
trade-off between precision and recall can provide an accurate indication of the overall
video saliency detection performance.
The F-measure can be computed via
\begin{equation}
\label{eq:FMeasure}
\mathrm{F}\text{-}\mathrm{measure} = \frac{(\beta^2+1)\times\mathrm{Precision}\times\mathrm{Recall}}{\beta^2\times\mathrm{Precision}+\mathrm{Recall}},
\end{equation}
where $\mathrm{Precision}$ is the average precision rate,
$\mathrm{Recall}$ is the average recall rate, and
$\beta^2=0.3$ is used to bias toward the precision rate, as was first suggested in~\cite{Contrast_FT:2009} and subsequently adopted by many significant studies.

MAE is defined as the average per-pixel difference between a saliency map $\textbf{S}$ and its corresponding ground truth $\textbf{G}$.
Here, $\textbf{S}$ and $\textbf{G}$ are normalized to the range [0,1].
\begin{equation}
\label{eq:MAE}
{\rm MAE} = \frac{\sum abs(\textbf{S}-\textbf{G})}{W \times H},
\end{equation}
where $W$ and $H$ are the width and height of the saliency map, respectively.

\subsection{Adaptiveness Analysis}
\label{sec:AdaptivenessAnalysis}
To validate the adaptiveness, we have tested our weakly supervised online learning scheme over five off-the-shelf color saliency deep models and one video saliency deep model, including RADF~\cite{hu2018recurrent}, ResDSS~\cite{CMMTPAMI2019resDSS}, RAS~\cite{chen2018eccvRAS}, CPD~\cite{wuCVPR2019cascaded}, PoolNet~\cite{liuCVPR2019simple} and SSAV~\cite{fanCVPR2019shifting}.
We have conducted the component evaluation to prove the effectiveness of our approach, with the results presented in Table~\ref{tab:NewDataComponent}.
Specifically, due to the use of our novel motion saliency, the fused low-level saliency exhibits a significant performance improvement.
Furthermore, the overall performance of our novel models also persistently and remarkably outperforms its low-level saliency.
In particular, as for the FBMS dataset, the color saliency deep models based solely on spatial information perform well for this dataset because the FBMS dataset is dominated by spatial information.
Nevertheless, benefiting from the newly sensed temporal information, our method still outperforms the color saliency deep models apparently according to the import metric maxF.

The experimental results show that the experimental model SSAV performs well on two datasets, but finetuning SSAV on the target video suffers from the performance decrease in terms of some metrics of some datasets.
Because the SSAV is a VIDEO saliency deep model, the only choice to incorporate the SSAV into our learning framework is to treat its saliency maps as the color saliency directly.
Actually, the effectiveness of our keyframe selection strategy is rooted in an assumption that the consistency degree between motion saliency and color saliency can well represent the quality of low-level saliency maps.
However, the saliency maps of SSAV usually are abundant in temporal information, which inevitably lead to persistent strong consistency between the so-called color saliency (i.e., the SSAV saliency maps) and the motion saliency, failing to select those really helpful keyframes.
Thus, the overall performance may get slightly worse after integrating the video method SSAV into our learning framework.

Moreover, though the proposed keyframe selection strategy can correctly select those frames with high-quality fused low-level saliency maps as the keyframes in the most cases (see Fig.~\ref{fig:KeyframeDistribution}), it may not always hold, as a result, there may exist exceptions occasionally that some video frames with low-quality low-level saliency maps get selected, leading to slight performance decrease (e.g., the experimental model CPD in terms of avgF of the UVSD~\cite{liu2017saliencySGSP} dataset).

In summary, the results of all of the quantitative experiments indicate that our method can adapt any off-the-shelf image saliency model for video data, achieving detection performance comparable to that of the state-of-the-art video saliency methods (e.g., SSAV19~\cite{fanCVPR2019shifting}).

\begin{table}[htbp]
	\vspace{-0.2cm}
	\caption{Ablation study toward the training iterations over the DAVIS and MCL datasets using the experimental RADF model. }
\resizebox{1\columnwidth}{!}{
		\begin{tabular}{c|c|c|c|c|c}
    \toprule[1pt]
			\multirow{2}[1]{*}{DataSet} & \multirow{2}[1]{*}{Metric}
			
&
			  \multicolumn{1}{c}{Iterations:} &
		\multicolumn{1}{c}{Iterations:} & \multicolumn{1}{c}{Iterations:} & Iterations:
		\\

			&    & \multicolumn{1}{c}{$(\lambda=1)\times N$}  &  \multicolumn{1}{c}{$(\lambda=5)\times N$}
			 &  \multicolumn{1}{c}{$(\lambda=8)\times N$} & \multicolumn{1}{c}{$(\lambda=10)\times N$}
			\\
			
			\midrule
			\multirow{3}[1]{*}{DAVIS~\cite{Perazzi2016benchmark}} & \multicolumn{1}{c|}{maxF} &  \multicolumn{1}{c}{{0.871}}  & \multicolumn{1}{c}{0.885} & \multicolumn{1}{c}{{0.885}} & {0.885}
			\\
			
			& \multicolumn{1}{c|}{avgF} & \multicolumn{1}{c}{{0.799}} & \multicolumn{1}{c}{0.815} & \multicolumn{1}{c}{0.815} & {0.813}
			\\

			& \multicolumn{1}{c|}{MAE} & \multicolumn{1}{c}{{0.032}} & \multicolumn{1}{c}{0.029} & \multicolumn{1}{c}{0.029} & {0.029}
			
			\\

			\midrule
			\multirow{3}[1]{*}{MCL~\cite{Kim:2015}} & \multicolumn{1}{c|}{maxF} &  \multicolumn{1}{c}{{0.768}}  & \multicolumn{1}{c}{0.792} & \multicolumn{1}{c}{{0.793}} & {0.793}
			\\
			
			& \multicolumn{1}{c|}{avgF} & \multicolumn{1}{c}{{0.655}} & \multicolumn{1}{c}{0.675} & \multicolumn{1}{c}{0.679} & {0.678}
			\\
			
			& \multicolumn{1}{c|}{MAE} & \multicolumn{1}{c}{{0.045}} & \multicolumn{1}{c}{0.041} & \multicolumn{1}{c}{0.041} & {0.041}
			
			\\

    \toprule[1pt]
		\end{tabular}%
	}
    \vspace{-0.4cm}
	\label{tab:NewDataMotion}%
\end{table}%

\begin{figure*}[!t]
	\begin{center}
		\includegraphics[width=1.0\linewidth]{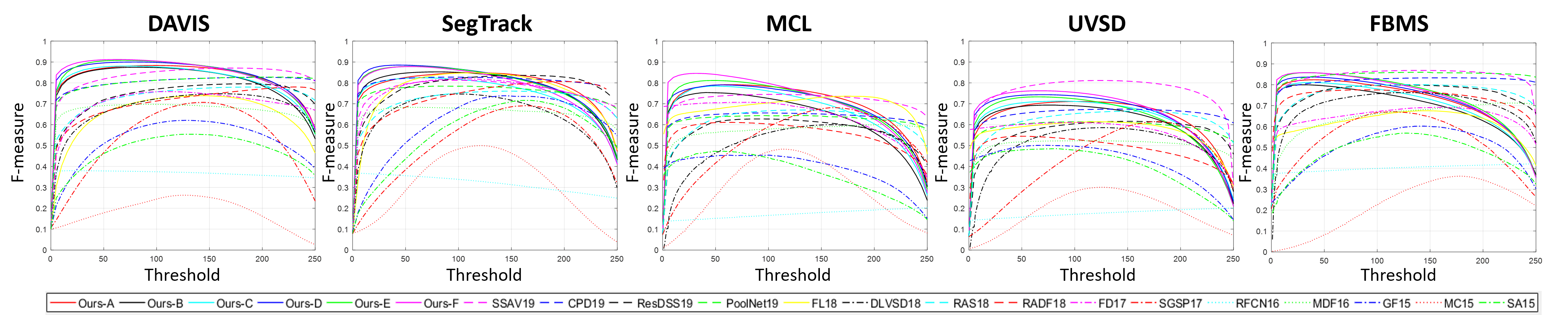}
	\end{center}
	\vspace{-0.4cm}
	\caption{Quantitative comparison (F-measure curves) between our 6 re-trained saliency deep models (5 image saliency models and 1 video saliency model) and 15 state-of-the-art methods over DAVIS2016(480p)~\cite{Perazzi2016benchmark}, SegV2~\cite{li2013video},
		MCL~\cite{Kim:2015},
		UVSD~\cite{liu2017saliencySGSP},
		and FBMS~\cite{ochs2014segmentation} datasets; the compared state-of-the-art methods include:
		SSAV19~\cite{fanCVPR2019shifting}
		CPD19~\cite{wuCVPR2019cascaded}, ResDSS19~\cite{CMMTPAMI2019resDSS}, PoolNet19~\cite{liuCVPR2019simple}, FL18~\cite{CC_TMM_2018},  DLVSD18~\cite{wang2018video}, RAS18~\cite{chen2018eccvRAS}, RADF18~\cite{hu2018recurrent}, FD17~\cite{TIP17chen2017video}, SGSP17~\cite{liu2017saliencySGSP}, RFCN16~\cite{wang2016saliency}, MDF16~\cite{TIP_li2016visual}, MC15~\cite{Kim:2015}, GF15~\cite{TIP15:2015}, SA15~\cite{wang2015saliency}.
		Ours-A,	Ours-B,	Ours-C, Ours-D, Ours-E, Ours-F denote the final saliency results after using our scheme over the model RADF, ResDSS, RAS, CPD, PoolNet, SSAV respectively.}
	\label{fig:Quantitative}
\end{figure*}

\subsection{Implementation Details}

We implement our method using Matlab2016b with the popular Caffe
platform.
All of the input frames are resized to the spatial resolution of 300$\times$300.
All of the quantitative evaluations are conducted
on a desktop computer with NVIDIA GTX 1080 GPU, Intel i7-6700k 4.00 GHz CPU (4 cores with 8 threads) and 32 GB RAM.
We conduct the training procedure using the widely adopted settings, namely, stochastic gradient descent (SGD) with a momentum of 0.95 and weight decay of 0.0005. We reduce the learning rate of the chosen image model by a factor of 0.1.
In our online training stage, assuming the number of the keyframe in the current video is $N$, the 6 tested saliency deep models (i.e., RADF, ResDSS, RAS, CPD, PoolNet, SSAV) were all trained by $\lambda\times N$ iterations, where the parameter $\lambda$ is empirically assigned to 8, and we have shown its ablation study in Table~\ref{tab:NewDataMotion}, in which the optimal choice of $\lambda$ can improve the overall performance by almost 1.5\% and 2.5\% in terms of both maxF and avgF in DAVIS~\cite{Perazzi2016benchmark} and MCL~\cite{Kim:2015} datasets, respectively.

As for the choice of the NR threshold mentioned in Sec.~\ref{sec:Low-level Saliency Guided Key Frame Selection}, we have conducted an ablation study on it over the challenging MCL~\cite{Kim:2015} dataset, see the quantitative proofs in Table~\ref{tab:NewDataNR}. On the one hand, since those frames of challenge video sequence usually have low-quality color or motion saliency leading to high NR values, it may be difficult to ensure a high diversity in those selected keyframes if we choose an extremely small NR.
On the other hand, a higher NR is more likely to result in more less-trustworthy keyframes. Moreover, the keyframe increase may also burden our online learning. Therefore, we decide to choose 0.6 as the threshold.

  \begin{table}[ht]
  	\centering
  	\vspace{-0.1cm}
  	\captionsetup{width=1\linewidth}
  	\caption{Ablation study on the non-overlapping ratio NR over the challenging MCL dataset using the experimental RADF model.}
  		\resizebox{1\linewidth}{!}{
  		\begin{tabular}{c|c|c|c|c|c|c}
    \toprule[1pt]
  			DataSet & Metric
  			&
  			\multicolumn{1}{c}{NR: 0.3} &
  			\multicolumn{1}{c}{NR: 0.5} & \multicolumn{1}{c}{NR: 0.6} &
  			\multicolumn{1}{c}{NR: 0.75} & NR: 0.85
  			\\
  			

  			\midrule
  			\multirow{3}[1]{*}{MCL~\cite{Kim:2015}} & \multicolumn{1}{c|}{maxF} &  \multicolumn{1}{c}{0.791}  & \multicolumn{1}{c}{0.792} & \multicolumn{1}{c}{0.793} & \multicolumn{1}{c}{0.789} & {0.791}
  			\\
  			
  			& \multicolumn{1}{c|}{avgF} & \multicolumn{1}{c}{{0.672}} & \multicolumn{1}{c}{0.680} & \multicolumn{1}{c}{0.679} & \multicolumn{1}{c}{0.673} & {0.670}
  			\\
  			
  			& \multicolumn{1}{c|}{MAE} & \multicolumn{1}{c}{{0.042}} & \multicolumn{1}{c}{0.042} & \multicolumn{1}{c}{0.041} & \multicolumn{1}{c}{0.043} & {0.042}
  			
  			\\

    \toprule[1pt]
  		\end{tabular}%
  	}
    \vspace{-0.2cm}
  	\label{tab:NewDataNR}%
  \end{table}%

For time analysis, since the keyframe number is determined by the total number of video frames in the given video sequence, our method may be somewhat time-consuming for adapting the color saliency deep model for video sequence with a large $N$; i.e., we have tested the average time per frame over all of the benchmarks. The runtime (second per frame) of all the methods are shown in Table~\ref{tab:TimeCost}.



\begin{figure*}[!t]
	\begin{center}
		\includegraphics[width=0.95\linewidth]{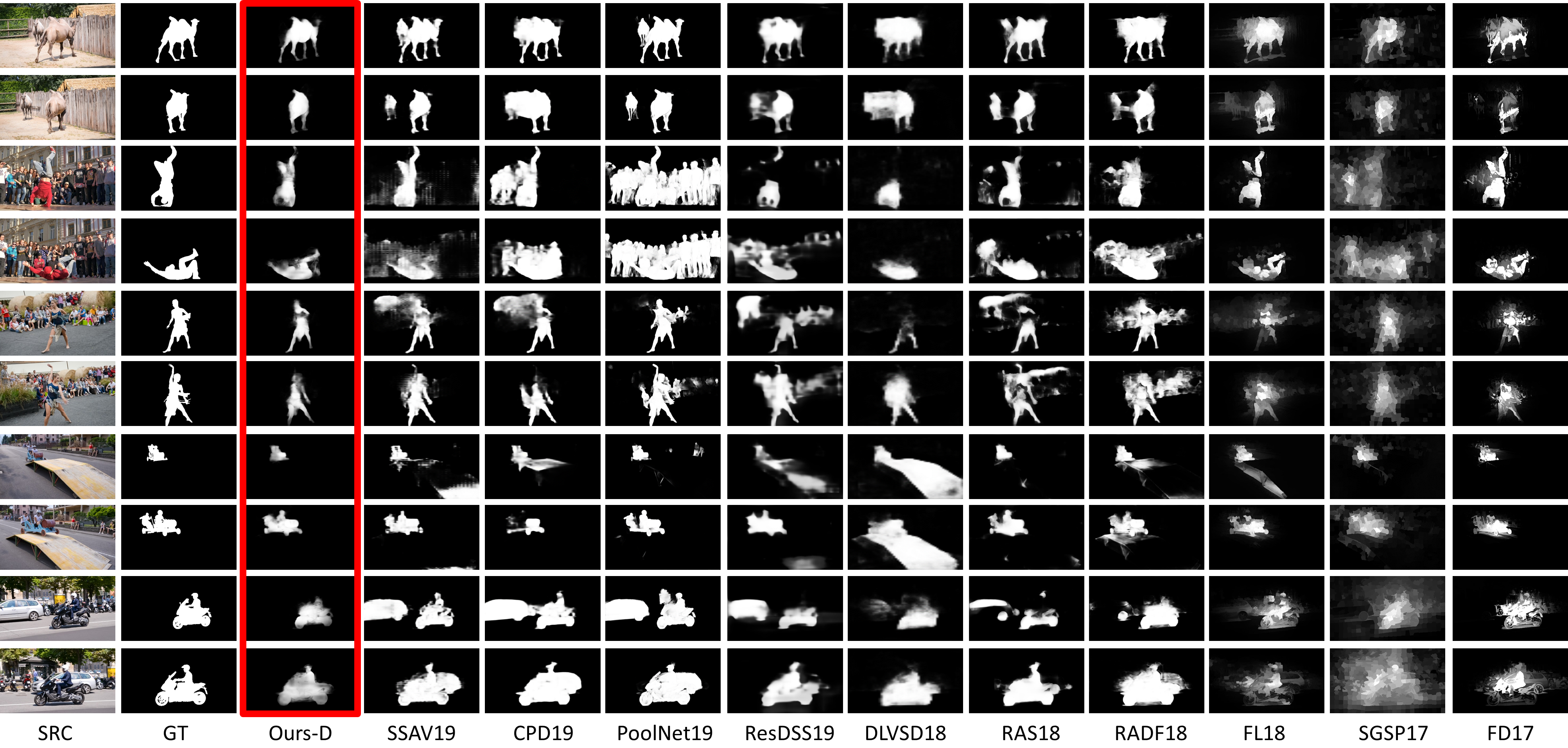}
	\end{center}
	\vspace{-0.4cm}
	\caption{Several most challenging sequences in our tested datasets. SRC denotes the source input video frames, GT shows the ground truth, Ours-D demonstrates the saliency maps obtained  by our experimental model CPD (highlighted with red rectangle), and column 4-13 demonstrates the results for some state-of-the-art methods, including:
		SSAV19~\cite{fanCVPR2019shifting},
		CPD19~\cite{wuCVPR2019cascaded}, PoolNet19~\cite{liuCVPR2019simple}, ResDSS19~\cite{CMMTPAMI2019resDSS},  DLVSD18~\cite{wang2018video}, RAS18~\cite{chen2018eccvRAS}, RADF18~\cite{hu2018recurrent},  FL18~\cite{CC_TMM_2018}, SGSP17~\cite{liu2017saliencySGSP}, and FD17~\cite{TIP17chen2017video}.}
	\vspace{-0.4cm}
	\label{fig:Demo1}
\end{figure*}

 \begin{table*}[t]
  	\caption{Comparison of time cost (in seconds) for single video frame between our method and other SOTA methods.}
  	\resizebox{1\linewidth}{!}{
  		\begin{tabular}{c|c|c|c|c|c|c|c|c|c|c|c|c|c|c|c|c|c|c|c|c|c}
    \toprule[1pt]
  			\multicolumn{1}{c|}{Method} & \multicolumn{1}{c|}{Ours-A}
  			
  			&
  			\multicolumn{1}{c|}{Ours-B} &
  			\multicolumn{1}{c|}{Ours-C} & \multicolumn{1}{c|}{Ours-D} &
  			\multicolumn{1}{c|}{Ours-E} &
  			\multicolumn{1}{c|}{Ours-F} &
  			\multicolumn{1}{c|}{SSAV} &
  			\multicolumn{1}{c|}{CPD} & \multicolumn{1}{c|}{ResDSS} &
  			\multicolumn{1}{c|}{PoolNet} &
  			\multicolumn{1}{c|}{FL} &
  			\multicolumn{1}{c|}{DLVSD} & \multicolumn{1}{c|}{RAS} &
  			\multicolumn{1}{c|}{RADF} &
  			\multicolumn{1}{c|}{FD} &
  			\multicolumn{1}{c|}{SGSP} & \multicolumn{1}{c|}{RFCN} &
  			\multicolumn{1}{c|}{MDF} &
  			\multicolumn{1}{c|}{GF} &
  			\multicolumn{1}{c|}{MC} & SA
  			
  			\\  				
  			
  			\midrule
  			\multicolumn{1}{c|}{Cost} & \multicolumn{1}{c|}{1.61} &  \multicolumn{1}{c|}{1.33}  & \multicolumn{1}{c|}{0.79} & \multicolumn{1}{c|}{1.53} &
  			\multicolumn{1}{c|}{1.49} &
  			\multicolumn{1}{c|}{1.50} &
  			
  			\multicolumn{1}{c|}{0.050} & \multicolumn{1}{c|}{0.029} & \multicolumn{1}{c|}{0.14} & \multicolumn{1}{c|}{0.042} & \multicolumn{1}{c|}{2.63} & \multicolumn{1}{c|}{0.47} & \multicolumn{1}{c|}{0.034} &
  			\multicolumn{1}{c|}{0.19} & \multicolumn{1}{c|}{119.4}&
  			\multicolumn{1}{c|}{51.7} & \multicolumn{1}{c|}{1.84}&
  			\multicolumn{1}{c|}{12.3} & \multicolumn{1}{c|}{53.7} &
  			\multicolumn{1}{c|}{18.3} & 45.4
  			\\
    \toprule[1pt]
  		\end{tabular}%
  	}
  	\label{tab:TimeCost}%
  	\vspace{-0.4cm}
  \end{table*}%

\subsection{Comparison With the State of the Art}

We have compared our method with 15 state-of-the-art methods, including SSAV19~\cite{fanCVPR2019shifting}, CPD19~\cite{wuCVPR2019cascaded}, ResDSS19~\cite{CMMTPAMI2019resDSS}, PoolNet19~\cite{liuCVPR2019simple}, FL18~\cite{CC_TMM_2018},  DLVSD18~\cite{wang2018video}, RAS18~\cite{chen2018eccvRAS}, RADF18~\cite{hu2018recurrent}, FD17~\cite{TIP17chen2017video}, SGSP17~\cite{liu2017saliencySGSP}, RFCN16~\cite{wang2016saliency}, MDF16~\cite{TIP_li2016visual}, MC15~\cite{Kim:2015}, GF15~\cite{TIP15:2015}, and SA15~\cite{wang2015saliency}.
The quantitative comparison results (the F-measure curves) are presented in Fig.~\ref{fig:Quantitative}.
As shown in Fig.~\ref{fig:Quantitative}, compared with SSAV19~\cite{fanCVPR2019shifting}, many of our newly adapted deep models achieve comparable detection performance.
As for other state-of-the-art methods, each of our newly adapted deep models, namely, RADF (Ours-A), ResDSS (Ours-B), RAS (Ours-C), CPD (Ours-D), PoolNet (Ours-E) and SSAV (Ours-F), significantly outperform all of them on DAVIS, SegV2, MCL and UVSD datasets.
Furthermore, the detailed maxF, avgF and MAE values can be found in Table~\ref{tab:NewData}, in which all the metric details suggest the superiority of our methods over the challenging DAVIS, SegV2, MCL and UVSD datasets.
In addition, our method has achieved the top-two best MAE score and avgF score in most of the tested datasets.

However, our method fails to perform well over the FBMS dataset, which is mainly due to the fact that the FBMS dataset is dominated by spatial information with frequent intermittent movement, making the video saliency detection by using both the spatial and temporal saliency cues much more difficult.
Benefiting from the long-term attribute of our method, our method still can achieve the top three maxF value for the FBMS dataset, as shown in Table~\ref{tab:NewData}.

We qualitatively compare the results of the different methods in Fig.~\ref{fig:Demo1}.
As shown in rows 1-2 of Fig.~\ref{fig:Demo1}, our method handles these long-period motionless sequences well.
Moreover, in such cases, almost all of the current state-of-the-art video saliency detection methods
easily give massive failure detections.
Furthermore, our methods can still handle the video scenes with complex backgrounds well; such video scenes are usually correlated with a challenging saliency estimation over the spatial domain, proving the effectiveness of our method for adapting the color saliency deep models for temporal saliency estimation, as shown in rows 3-10 of Fig.~\ref{fig:Demo1}.

The quantitative results obtained by the most recent advanced video saliency detection method SSAV~\cite{fanCVPR2019shifting} are shown in Table~\ref{tab:NewData}, and it is observed that the performance of our method is comparable to that of SSAV.
Specifically, Ours-F clearly outperforms SSAV by 4.1\%, 6.5\%, and 10.1\% in terms of maxF over the DAVIS, SegV2 and MCL datasets, respectively.
Furthermore, Ours-F achieved an MAE values that is very close to the MAE for the SSAV method on DAVIS and SegV2.
It should also be noted that the SSAV method adopts an extremely large training dataset (with the additional eye fixation data), while by contrast, our deep models are trained using the Davis training set.
With the rapid development of the color saliency deep models, we believe that our method will eventually outperform the SSAV method.

\vspace{0.2cm}
\section{Conclusion}

This paper has proposed a novel weakly supervised scheme to adapt image saliency deep models for video data. Our method can generate a novel motion saliency sub-branch via fine-tuning the off-the-shelf image saliency deep model using the color-coded optical flow data.
Furthermore, we propose the newly-designed keyframe strategy to locate the frames with high-quality spatiotemporal saliency predictions.
Then, we have used these high-quality predictions as the pseudo ground truth for the weakly supervised online training, enabling all of the off-the-shelf image saliency deep models to be adapted for the current video sequence as the new color sub-branch of our method.
Our method is simple, flexible, and effective, and is likely to inspire future work even in the case that our color model adapted method is only comparable to the current leading state-of-the-art video saliency detection methods.

\textbf{Acknowledgments}. This research was supported in part by National Key R\&D Program of China (No. 2017YFF0106407), National Natural Science Foundation of China (No. 61802215 and No. 61806106), Natural Science Foundation of Shandong Province (No. ZR201807120086) and National Science Foundation of USA (No. IIS-1715985, IIS0949467, IIS-1047715, and IIS-1049448).
\vspace{-0.2cm}

\bibliographystyle{IEEEtran}
\bibliography{egbibsample}

\end{document}